\documentclass[times,twocolumn,final]{elsarticle}
\usepackage{medima}
\usepackage{framed,multirow}
 \usepackage{booktabs}
\usepackage{amssymb}
\usepackage{latexsym}
\usepackage{tabularx}
\usepackage{colortbl} 
\usepackage{url}
\usepackage{xcolor}
\usepackage{hyperref}
\usepackage{xspace}
\usepackage{amsmath,amssymb,amsfonts}
\usepackage[caption=false,font=scriptsize,labelfont=rm,textfont=rm]{subfig}
\usepackage[ruled,vlined,linesnumbered,noend]{algorithm2e}
\usepackage{graphicx}

\newcommand{\ie}{\textit{i.e.}\xspace}
\newcommand{\eg}{\textit{e.g.}\xspace}
\newcommand{\etal}{\textit{et al.}\xspace}

\newcommand{\resp}{\textit{resp.}\xspace}

\definecolor{newcolor}{rgb}{.8,.349,.1}

\journal{Medical Image Analysis}

\begin{document}

\verso{Y. Gu \textit{et~al.}}

\begin{frontmatter}

\title{Dual Structure-Aware Image Filterings for Semi-supervised Medical Image Segmentation}%
% \tnotetext[tnote1]{This is an example for title footnote coding.}

% \author[1]{Yuliang \snm{Gu}\corref{cor1}}
% \cortext[cor1]{Corresponding author: 
%   Tel.: +0-000-000-0000;  
%   fax: +0-000-000-0000;}
% \author[1]{Given-name2 \snm{Surname2}\fnref{fn1}}
% \fntext[fn1]{This is author footnote for second author.}
% \author[2]{Given-name3 \snm{Surname3}}
% %% Third author's email
% \ead{author3@author.com}
% \author[2]{Given-name4 \snm{Surname4}}

% \address[1]{Affiliation 1, Address, City and Postal Code, Country}
% \address[2]{Affiliation 2, Address, City and Postal Code, Country}

\author[1,2,3]{Yuliang \snm{Gu}}
\ead{yuliang_gu@whu.edu.cn}
\author[1,2,3]{Zhichao \snm{Sun}}
\ead{zhichaosun@whu.edu.cn}
\author[1,2,3]{Tian \snm{Chen}}
\ead{tian.chen@whu.edu.cn}
\author[1,2,3]{Xin \snm{Xiao}}
\ead{xinxiao@whu.edu.cn}
\author[1,2,3]{Yepeng \snm{Liu}}
\ead{yepeng.liu@whu.edu.cn}
\author[1,2,3]{Yongchao \snm{Xu} \corref{cor1}}
\cortext[cor1]{Corresponding author}
\ead{yongchao.xu@whu.edu.cn}
\author[4]{Laurent \snm{Najman}}
\ead{laurent.najman@esiee.fr}

\address[1]{National Engineering Research Center for Multimedia Software, Wuhan University, Wuhan, China}
\address[2]{Institute of Artificial Intelligence, School of Computer Science, Wuhan University, Wuhan, China}
\address[3]{Hubei Key Laboratory of Multimedia and Network Communication Engineering, Wuhan University, Wuhan, China}
\address[4]{Univ Gustave Eiffel, CNRS, LIGM, Marne-la-Vall$\acute{e}$e, France}

% \received{1 May 2013}
% \finalform{10 May 2013}
% \accepted{13 May 2013}
% \availableonline{15 May 2013}
% \communicated{S. Sarkar}

\begin{abstract}
Semi-supervised image segmentation has attracted great attention recently. 
The key is how to leverage unlabeled images in the training process. 
Most methods maintain consistent predictions of the unlabeled images under variations (\eg, adding noise/perturbations, or creating alternative versions) in the image and/or model level. 
%These variations are usually achieved via addition of noise, perturbations, or the creation of alternative versions of the image or model. 
% In particular, image-level variations have been proved useful in semi-supervised natural image segmentation. 
% Yet, medical images often have prior structure information, which has not been well explored in most image-level variations. 
In most image-level variation, medical images often have prior structure information, which has not been well explored.
In this paper, we propose novel dual structure-aware image filterings (DSAIF) as the image-level variations for semi-supervised medical image segmentation. 
Motivated by connected filtering that simplifies image via filtering in structure-aware tree-based image representation, we resort to the dual contrast invariant Max-tree and Min-tree representation.
%given by the inclusion relationship between connected components of the upper and lower level sets, respectively. 
Specifically, we propose a novel connected filtering that removes topologically equivalent nodes (\ie connected components) having no siblings in the Max/Min-tree. This results in two filtered images preserving topologically critical structure. 
Applying the proposed DSAIF to mutually supervised networks decreases the consensus of their erroneous predictions on unlabeled images. This helps to alleviate the confirmation bias issue of overfitting to noisy pseudo labels of unlabeled images, and thus effectively improves the segmentation performance. 
Extensive experimental results on three benchmark datasets demonstrate that the proposed method significantly/consistently outperforms some state-of-the-art methods. The source codes will be publicly available.
\end{abstract}

% Specifically, we resort to the dual contrast invariant and invertible tree-based image representations: Max-tree and Min-tree given by the inclusion relationship between connected components of the upper and lower level sets, respectively. For each successive set of topologically equivalent nodes (\ie connected components) having no siblings in the tree, we remove these nodes and set the closest ancestor node as the parent of their children nodes. This results in simplified Max-tree and Min-tree, from which we reconstruct two structure-preserving filtered images. We then apply mutual supervision on these two filtered images for semi-supervised medical image segmentation.

% based image transformation methods, we leverage a tree-based image representation to generate diverse augmented views for the same image. We design small area pruning algorithm and component-preserving algorithm in Max-tree and Min-tree space to generate structure and shape related diversity transformation, which effectively utilizes the shape prior information of the organs to be segmented. This approach effectively harnesses structured information encompassing grayscale values and geometric shapes. 

\end{frontmatter}

%\linenumbers

\section{Introduction}
\label{sec:introduction}
Accurate medical image segmentation plays an important role in computer-aided diagnosis (CAD) systems. Traditional supervised segmentation methods have achieved impressive results using a large amount of labeled data. Yet, the manual segmentation is laborious and time-consuming. Recently, semi-supervised segmentation methods have gained significant attention by utilizing easily accessible unlabeled images to improve the accuracy of segmentation models.

% Most existing semi-supervised segmentation methods can be roughly divided into two categories: consistency regularization methods and training with pseudo labels. The former are mainly based on smoothness assumption~\citep{chen2022semi} which aims to produce consistent  results under small perturbations at data level and model level. 

The mainstream semi-supervised segmentation methods are based on consistency regularization~\citep{zhao2023augmentation, wang2023mcf,yang2023revisiting,basak2023pseudo, lei2022semi, jin2022semi,xiang2022fussnet,basak2022addressing, lyu2022pseudo, su2024mutual, adiga2024anatomically}, which aims to produce consistent results under variations at image-level or/and model-level. In particular, many approaches aim to generate variations under image-level~\citep{yu2019uncertainty, xu2021shadow, you2022momentum, you2022simcvd, bai2023bidirectional}. A popular strategy for image variations utilizes the weak-to-strong paradigm~\citep{fan2022ucc,liu2022perturbed,yang2023revisiting}, where predictions generated from weakly-augmented versions are used to supervise the strongly-augmented versions. Augmented versions are usually generated by simple random augmentation (\eg, Gaussian noise~\citep{huang2022semi}), adversarial perturbation~\citep{peiris2021duo, wang2023cat}, and CutMix techniques~\citep{chen2021semi,yang2023revisiting}. The model-level variations mainly adopt the Mean Teacher framework~\citep{tarvainen2017mean} or Co-training strategy~\citep{qiao2018deepcotraining, chen2021semi}. In the Mean Teacher framework, the teacher network is usually obtained from the student network via Exponential Moving Average (EMA). The co-training strategy involves training two independent networks or decoders with different initializations and using each model's outputs to supervise the other's training in a mutual fashion. 

Recently, the consistency regularization methods using pseudo labels for supervision have achieved impressive performance for semi-supervised segmentation~\citep{chen2021semi, yang2023revisiting, basak2023pseudo, lyu2022pseudo, liu2022semi}. For instance,
CPS~\citep{chen2021semi} generates different pseudo labels by two networks with different initializations and applies mutual supervision between them. These methods have achieved impressive performance in natural images, thanks to effective strong image augmentation (\eg, CutMix~\citep{yun2019cutmix}) as image-level variations for avoiding the model overfit to incorrect pseudo-labels~\citep{chen2021semi, yang2023revisiting, liu2022perturbed}. However, these existing image-level variations do not make well use of the structure information, which is important for medical images. Moreover, the distribution variance in medical images is not as significant as in natural images, which makes the semi-supervised medical image segmentation more prone to overfit noisy pseudo-labels due to confirmation bias~\citep{arazo2020pseudo}.
% However, objects of interest in medical images usually have some shape priors. Existing CutMix-like data augmentation methods do not consider such prior structure information. 

In this paper, we propose novel dual structure-aware image filterings (DSAIF), serving as the image-level variations to cope with the confirmation bias in semi-supervised medical image segmentation. For that, we aim to obtain two filtered images with diverse image appearances, while preserving the critical topological structure of the original image. Specifically, we resort to the dual contrast-invariant Max-tree and Min-tree~\citep{salembier1998antiextensive} representation, given by the inclusion relationship between connected components of upper and lower level sets, respectively. The topology of the tree structure encodes the topology of the image structure. Such structure-aware tree-based image representation is widely used to implement connected filterings~\citep{salembier1998antiextensive,westenberg2007volumetric, wilkinson2008concurrent, ouzounis2007mask, xu2015connected} that do not create new edges. We propose a novel type of connected filtering that preserves the topological structure of image. Precisely, we remove all nodes (\ie connected components) having no siblings in the Max-tree and Min-tree, resulting in two simplified trees preserving topologically critical structure. The corresponding filters named upper/lower structure-aware image filtering (denoted as USAIF and LSAIF) give rise to two different images having the same topological structure as the original image. %An illustrative example is given in Fig.~\ref{fig:image2tree}. 

To further cope with the confirmation bias issue on unlabeled medical images, we also propose to apply monotonically increasing contrast changes before performing the dual structure-aware image filterings. Since the Max-tree and Min-tree are invariant to such increasing changes, the resulting filtered images still preserve the topological image structure while having large diversity in image appearances. 
By incorporating the proposed DSAIF into mutually supervised networks, the consensus on incorrect predictions for unlabeled images is decreased. This helps to alleviate the confirmation bias issue, where models tend to overfit to noisy pseudo labels, thereby enhancing the performance of segmentation.
Applying such dual structure-aware image filterings as the image-level variations decreases the consensus of erroneous predictions for unlabeled images. This helps to alleviate the confirmation bias issue of overfitting to noisy pseudo labels of unlabeled images, thereby enhancing the performance for semi-supervised medical image segmentation.
We adopt the mutual supervision framework of CPS~\citep{chen2021semi} and MC-Net~\citep{wu2021semi} as the baseline models. The proposed DSAIF significantly boosts the performance of CPS and MC-Net baseline, and significantly/consistently outperforms some state-of-the-art
methods on three benchmark datasets.

The main contribution of the paper is summarized as follows: 
1) We propose novel dual structure-aware image filterings (DSAIF) as the image-level variations for semi-supervised medical image segmentation. DSAIF yields two images with quite different appearances while having the same topological structure as the original image.  
2) We further leverage the contrast-invariance property of Max/Min-tree representation involved in DSAIF. We apply monotonically increasing contrast changes before performing DSAIF. This increases the appearance diversity while preserving topological image structure.
3) The proposed method significantly/consistently outperforms some state-of-the-art methods on three widely benchmark datasets. In particular, using only 20\% of labeled images, the proposed method achieves similar ($\sim$99.5\%) segmentation performance with the use of full dataset.

The rest of this paper is organized as follows. We first review some related works in Section~\ref{sec:related-work}, followed by the detail of the proposed method in Section~\ref{sec:method}. We then present extensive experimental results in Section~\ref{sec:experiments}. Finally, we conclude in Section~\ref{sec:conclusion}.
\section{Related work}
\label{sec:related-work}

\begin{figure*}[!t]
\centering
\includegraphics[width=7in]{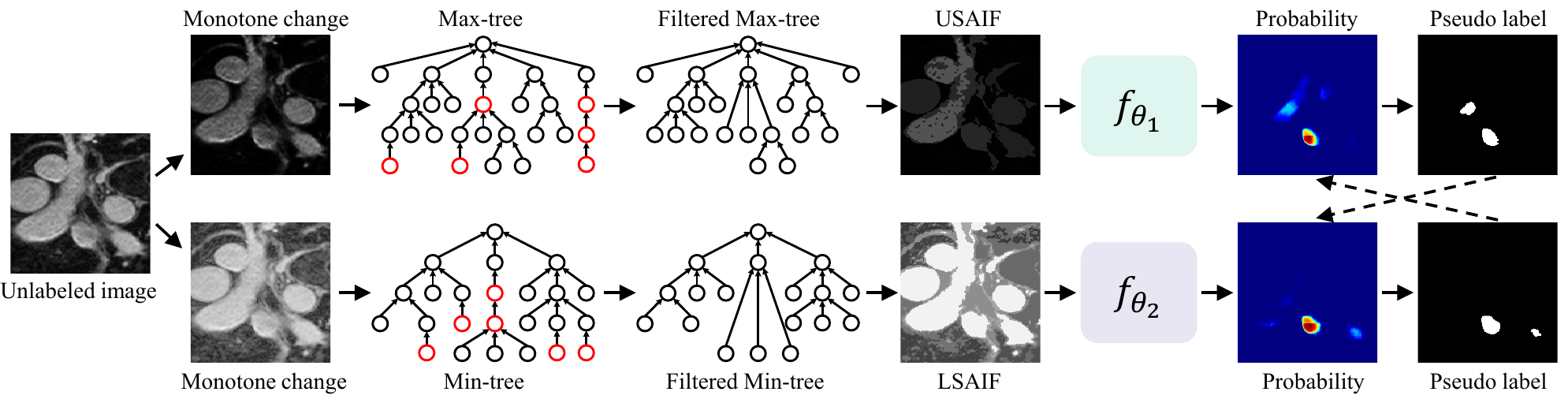}
\caption{The pipeline of the proposed DSAIF framework using mutual supervision of CPS~\cite{chen2021semi} as the model-level variations. We propose novel dual structure-aware image filterings (DSAIF) based on Max/Min-tree representation as the image-level variations. We remove every node (marked in red) without siblings in Max/Min-tree which is topologically equivalent to its ancestor node. } 
% \caption{The pipeline of the proposed DSAIF framework using mutual supervision of CPS~\cite{chen2021semi} as the model-level variations. The pipeline composed of image-level variations and model-level variations on images. We propose novel dual structure-aware image filterings (DSAIF) based on Max/Min-tree representation as the image-level variations.} 
\label{fig:pipeline}
\end{figure*}

\subsection{Semi-supervised Learning}
Semi-supervised learning (SSL) aims to leverage limited annotated data and a large number of unlabeled data to improve the performance. Existing semi-supervised learning methods can be roughly grouped into two categories~\citep{chen2022semi}: self-training and consistency regularization. Self-training methods~\citep{grandvalet2005semi} learn from unlabeled data by assigning pseudo labels to unlabeled data and subsequently integrating them with manually labeled data for further retraining. Consistency regularization methods~\citep{laine2016temporal, tarvainen2017mean} are mainly based on smoothness assumption~\citep{chen2022semi} which aims to produce consistent  results under small variations at image-level and/or model-level.  The simple random augmentation~\citep{sajjadi2016regularization} and adversarial perturbation~\citep{ miyato2018virtual} are representative works of image-level perturbations for consistency regularization. There are mainly three types of model-level perturbations: 1) Directly adding  stochastic perturbation (\eg, Gaussian noise~\citep{rasmus2015semi} or dropout~\citep{park2018adversarial}) to the model weights;
2) Mean Teacher~\citep{tarvainen2017mean} that ensembles model’s parameters produced during training using exponential moving average (EMA) strategy;
3) Generating model variations via different decoders~\citep{wu2022mutual} or networks~\citep{qiao2018deepcotraining}.

\subsection{Semi-supervised Medical Semantic Segmentation}
Semi-supervised learning is widely used in  medical image segmentation tasks thanks to its ability in alleviating the difficulty of manually annotating medical images. 

Methods~\citep{basak2023pseudo, lei2022semi, jin2022semi, xiang2022fussnet, basak2022addressing, wang2023mcf, lyu2022pseudo,adiga2024anatomically, su2024mutual} based on consistency regularization have achieved impressive performance for semi-supervised medical semantic segmentation. These methods usually use Mean Teacher framework~\citep{tarvainen2017mean} or Co-training strategy~\citep{qiao2018deepcotraining} to generate variations in the model-level. Another way aims to generate diverse versions of the same image and  enforce prediction consistency under image variations~\citep{huang2022semi, xu2021shadow, fan2022ucc, peiris2021duo, wang2023cat}. A typical approach for image variations involves the weak-to-strong paradigm~\citep{fan2022ucc}, where weakly-augmented and strongly-augmented images are employed to promote consistency. Methods~\citep{peiris2021duo, wang2023cat} incorporate adversarial training strategy to generate adversarial perturbations on images and make the predictions robust to adversarial perturbations. Recently, an increasing number of methods enhance model performance by training unlabeled images with pseudo labels~\citep{lyu2022pseudo, qiao2022semi}. Since there are inevitable noisy labels in the pseudo labels for unlabeled images, it is crucial to determine the confidence level of pseudo-labels~\citep{qiao2022semi, wang2021semi}. Moreover, some methods~\citep{liu2022perturbed} focus on pseudo rectifying during the training stage. Apart from these approaches, some methods~\citep{you2022simcvd, basak2023pseudo} exploit contrastive learning to achieve consistent feature representation.

Considering that objects of interest in medical images usually have specific shapes, some works~\citep{li2020shape,luo2021semi,meng2022dual, wang2021tripled, liu2022semi} also incorporate shape information to alleviate the problem of insufficient labeled images in semi-supervised medical image segmentation. For instance, Li \etal~\citep{ li2020shape} leverage signed distance map (SDM) of object surfaces as a multi-task predictiong jointly with semantic segmentation , and use an adversarial loss calculated by SDM as a geometric shape consistency constraint.  A dual-task network is used in~\citep{luo2021semi} to jointly predict segmentation maps and level set representations that can capture global-level shape and geometric information of the target. Wang \etal~\citep{wang2021tripled} extend the mean teacher architecture with foreground and background reconstruction task and signed distance field  prediction task to combine semantic information and shape information. 
% However, existing shape-related methods tend to consider contour information solely as a representation of shape, utilizing edge distance functions to describe its characteristics. 
%However, these shape-related methods simply consider contour as shape information and employ edge distance functions to characterize the shape. Besides the contour, the definition of shape holds rich connotations. In this article, we explore the utilization of tree-based images to represent the comprehensive shape information of organs in medical images.

\subsection{Tree-based image representation}
Typically, an image is usually modeled as a discrete function defined on pixels or voxels over a 2D or 3D domain $V (\mathbb{R}^{2} \text{or}~ \mathbb{R}^{3})$. However, in the field of image processing and computer vision, many applications rely on interacting with some primitives of fundamental elements being more meaningful than the pixels. The tree-based image representation~\citep{xu2014tree, xu2015connected, xu2016hierarchical} is composed of a set of regions of the original image. These regions are either disjoint or have inclusion relationship between them, and thus can be encoded into a tree structure. Hierarchical segmentation and threshold decomposition are two main branches of tree-based image representations.  A hierarchy of segmentation consists of a set of fine to coarse partitions.  
This hierarchy can be depicted as a tree structure, with the root node representing the entire image as a unified region, and the leaf nodes denoting the regions within the finest image partition. The intermediate nodes, situated between the root and the leaves, represent regions obtained through the fusion of all the regions represented by their child nodes.  The $\alpha$-tree~\citep{soille2008constrained} and the binary partition tree (BPT)~\citep{salembier2000binary} are two popular works of hierarchical segmentation. 
%Those hierarchies trees are widely used in image segmentation\citep{xu2016hierarchical}, image simplification\citep{xu2016hierarchicaimagel}, and connected filtering\citep{xu2015connected}.
\begin{figure*}[t]
\centering
\includegraphics[width=7in]{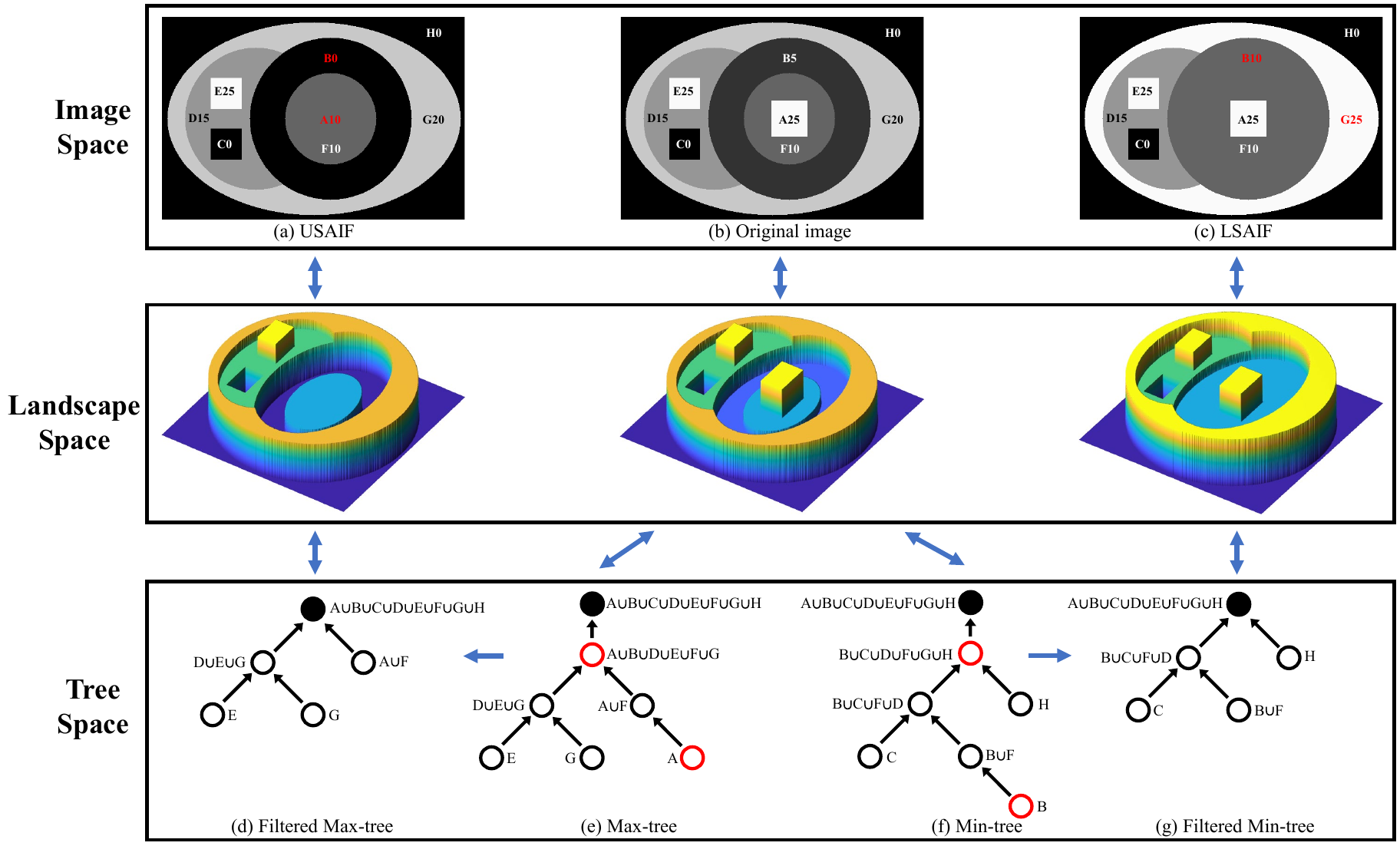}
\caption{An illustrative example of the proposed DSAIF. For the Max-tree and Min-tree built on the original image (b), we remove every node (marked in red) without siblings which is topologically equivalent to its ancestor node. The two images reconstructed from filtered Max/Min-tree denoted as USAIF (a) and LSAIF (c) have the same topological structure as the original image, but are of quite different appearances. The number after the letter denotes the graylevel of the region.}
%\caption{A synthetic image that explain the tree-based image representation and the component-preserving filter generating structure and shape related diverse image views. (e) is the Max-tree representation of original synthetic image (b), and (f) is the corresponding Min-tree representation. The critical regions are represented by red circles which has only one child, and the solid circles represent root nodes. After component-preserving filtering, the Max-tree is represented as (d), and the Min-tree is  represented as (g). Transform the filtered Max-tree into a pixel based image representation as shown in (a), and the same with Min-tree as shown in (c). }
\label{fig:image2tree}
\end{figure*}

Threshold decompositions developed in mathematical morphology are another widely used type of tree-based image representation. Image representations based on threshold decomposition rely solely on pixel-value ordering, rendering the generated tree structures invariant to monotonically increasing contrast changes. Embedding the set of upper level sets into a tree structure gives the Max-tree~\citep{salembier1998antiextensive}. The root of Max-tree represents the entire image domain, and the leaves correspond to the local regional maxima of the image. By duality, the lower level sets give rise to Min-tree representation~\citep{salembier1998antiextensive}. The root of Min-tree also represents the entire image domain, while the leaves correspond to the local regional minima of the image. The Max/Min-tree can be computed with quasi-linear complexity based on Union-Find process~\citep{najman2006building, carlinet2014comparative}. Topographic map~\citep{caselles1999topographic}, also known as tree of shapes~\citep{monasse2000fast} is another tree representation based on the threshold decomposition. It is derived by leveraging the  inclusion relationship of the shapes, where a shape is defined as the connected component of upper or lower level sets with holes filled. 
The tree structures constructed through threshold decomposition are all contrast-invariant, offering a multi-scale representation comprising a series of included or disjoint regions ranging from small to large scales~\citep{xu2014tree, xu2015connected}. These trees are proved to be useful in many applications, such as lymphoma tumor segmentation from PET imaging~\cite{grossiord2020shaping}, local feature detection~\citep{xu2014tree},  or classification of high resolution satellite images~\citep{luo2013robust}.

\section{Method}
\label{sec:method}

\subsection{Overview}
Semi-supervised semantic segmentation task aims to enhance the performance of segmentation by leveraging a small set of labeled images  $\mathcal{D}^l = \{(x^l, y^l)\}$ of $N$ labeled images, along with a large collection of unlabeled images  $\mathcal{D}^u = \{x^u\}$  of $M$ unlabeled images, where $N\ll M$. 

We follow classical consistency regularization-based semi-supervised medical image segmentation framework, which is often composed of image-level variations and model-level variations on unlabeled images. For the image-level variations, we resort to dual contrast-invariant Max-tree and Min-tree representation (see Sec.~\ref{subsec:tree-construction} for the construction) for connected filterings. We propose novel dual structure-aware image filterings (DSAIF) as the image-level variations. More specifically, we propose a novel type of connected filtering that preserves only the topologically critical nodes of Max/Min-tree. The corresponding filtering named upper/lower structure-aware image filtering denoted as USAIF/LSAIF, yields two different images that have the same topological structure as the original one. We further leverage the invariance property of Max/Min-tree with respect to monotonically increasing contrast changes to further enforce the appearance diversity while preserving the topological image structure. 
For the model variations, we simply adopt cross pseudo supervision (CPS) method~\citep{chen2021semi} as a baseline example to illustrate our method in Fig.~\ref{fig:pipeline}. 
It is noteworthy that DSAIF can also be applied to other mutual supervision framework such as MC-Net~\citep{wu2021semi}. The pipeline of the proposed framework using MC-Net as baseline is depicted in \textbf{Supplementary}.

\subsection{Tree Construction}
\label{subsec:tree-construction}

% The pixel-based image representation loses a lot of structural information of organs in medical image.
We utilize image threshold decompositions to build Max/Min-tree representation. By performing thresholding on a grayscale image $x$ in descending order, starting from $h_{max}$ to $h_{min}$, a sequence of nested upper level sets is obtained. Each upper level set at level $h$ is a binary image given by $\mathcal{X}_{h}(x)=\{v \in V | x(v) \geq h\}$. Let $ P_{h}^{v}(x) $ represents the binary connected operator of $\mathcal{X}_{h}(x)$ at point $v$, which gives the connected component of $\mathcal{X}_{h}(x)$ containing $v$ if $v \in \mathcal{X}_{h}(x)$, and $ \emptyset $ otherwise. Then, for any two connected components $P_{h_{1}}^{v_{1}}(x)$ and $P_{h_{2}}^{v_{2}}(x)$ at respectively level $h_{1} \geq h_{2} $, we have either $P_{h_{1}}^{v_{1}}(x) \subseteq P_{h_{2}}^{v_{2}}(x)$, or $P_{h_{1}}^{v_{1}}(x) \cap P_{h_{2}}^{v_{2}}(x) = \emptyset$. Based on this inclusion relationship, a tree structure named Max-tree is formed, where nodes correspond to connected components. The parenthood between nodes corresponds to the inclusion relationship between the underlying connected components. 

We use a water-covered surface analogy to better illustrate the process of Max-tree construction and the associated alterations in the level sets. For that, we suppose the surface is entirely submerged in water. With the level of water gradually decreasing, islands (regional maxima) emerge first to form the leaves of the tree.  As the water level continues to drop, these islands expand, building the tree's branches. At certain levels, multiple islands fuse into a single connected piece, creating forks (i.e., the nodes of the tree with several children) in the tree structure. This process continues until all the water has evaporated, leaving behind a solitary landmass which forms the tree's root, representing the entirety of the image. By duality, a corresponding dual structure of the Max-tree, known as the Min-tree, is constructed based on the decomposition of lower level sets defined by $\mathcal{X}^{h}(x)=\{v \in V | x(v) \leq h\}$. A synthetic example of Max-tree and Min-tree is given in Fig.~\ref{fig:image2tree}. The Max/Min-tree can be constructed efficiently using Union-Find-based algorithms~\citep{najman2006building, carlinet2014comparative}, which has a quasi-linear complexity with respect to the number of pixels. %More details and examples of tree construction can be found in the \textbf{Supplementary}. 

%Fig.~\ref{fig:tree transform example} displays an example of synthetic image (b), along with its corresponding Max-tree (e) and Min-tree (f). 

\subsection{Dual Structure-Aware Image Filterings}
\label{subsec:tree-filter}
The Max/Min-tree representation is equivalent to the original image in the sense that the image $x$ can be reconstructed from the tree $\mathcal{T}$, composed of a set of nodes $\{\mathcal{N}\}$ with inclusion relationship encoded by $parent$. Specifically, we associate the graylevel $h$ to the corresponding node on which the underlying connected component is obtained. Then, for each pixel $v \in V$, the grayscale value $x(v)$ is given by the associated graylevel of the smallest node containing $v$. Removing nodes from the tree and updating the corresponding parenthood relationship results in a simplified tree, from which a filtered image is reconstructed. This is one of the most popular implementations of connected filters.

\begin{algorithm}[t]
    \SetAlgoLined 
	\caption{Structure-aware image filtering. Small regions with area less than $\tau$ may be caused by noise, and do not contribute to the topological changes.}
        \label{alg:tree-component-preserving}        %\tcp{$P(n)$ : the parent node of $n$ }
        %\tcp{$C(n)$ : the child nodes of $n$ }
        %\tcp{$M(n)$ : the members of $n$ }
        % \tcp{$R(n)$ : the regions of $n$ }
        %\tcp{$area(n)$ : the number of pixel in regions of $n$ }
	%\KwIn{Max-tree or Min-tree $\mathcal{T} : P,C,M$.}
	%\KwOut{Filtered tree $\widetilde{\mathcal{T}}: \widetilde{P},\widetilde{C},\widetilde{M}$.}

        \SetKwFunction{FAREA}{STRUCT\_AWARE\_FILTER}
        \SetKwProg{Fn}{Function}{}{}
        \Fn{\FAREA{$x$, $\tau$}}{
            %\Begin{ 
            % $\widetilde{\mathcal{T}} = \mathcal{T}$\;
            $\mathcal{T}$ $\leftarrow$ $Compute\_Tree$($x$)\\
            \ForEach{$\mathcal{N} \in \mathcal{T}$}{$numChildren(\mathcal{N}) \leftarrow 0$\\ $isRemoved(\mathcal{N}) \leftarrow$ \textbf{False}
            }
            \ForEach{$\mathcal{N} \in \mathcal{T}$}{
            \eIf{$area(\mathcal{N}) > \tau$}{$++numChildren(parent(\mathcal{N}))$}
            {$isRemoved(\mathcal{N}) \leftarrow$ \textbf{True}}
            }
            \ForEach{$\mathcal{N} \in \mathcal{T}$}{
            \lIf{$numChildren(parent(\mathcal{N})) = 1$}{$isRemoved(\mathcal{N}) \leftarrow$ \textbf{True}}
            }
            \ForEach{$v \in V$}{
            $\mathcal{N} \leftarrow Get\_Node(v)$ //Smallest node contains $v$\\
            \While{ $isRemoved(\mathcal{N})$}{$\mathcal{N} \leftarrow parent(\mathcal{N})$}
            $x'(v) \leftarrow x(\mathcal{N})$ 
            }
            \textbf{return} $x'$
            }
         %}
\end{algorithm}

\begin{figure*}[!t]
\centering
\includegraphics[width=7in]{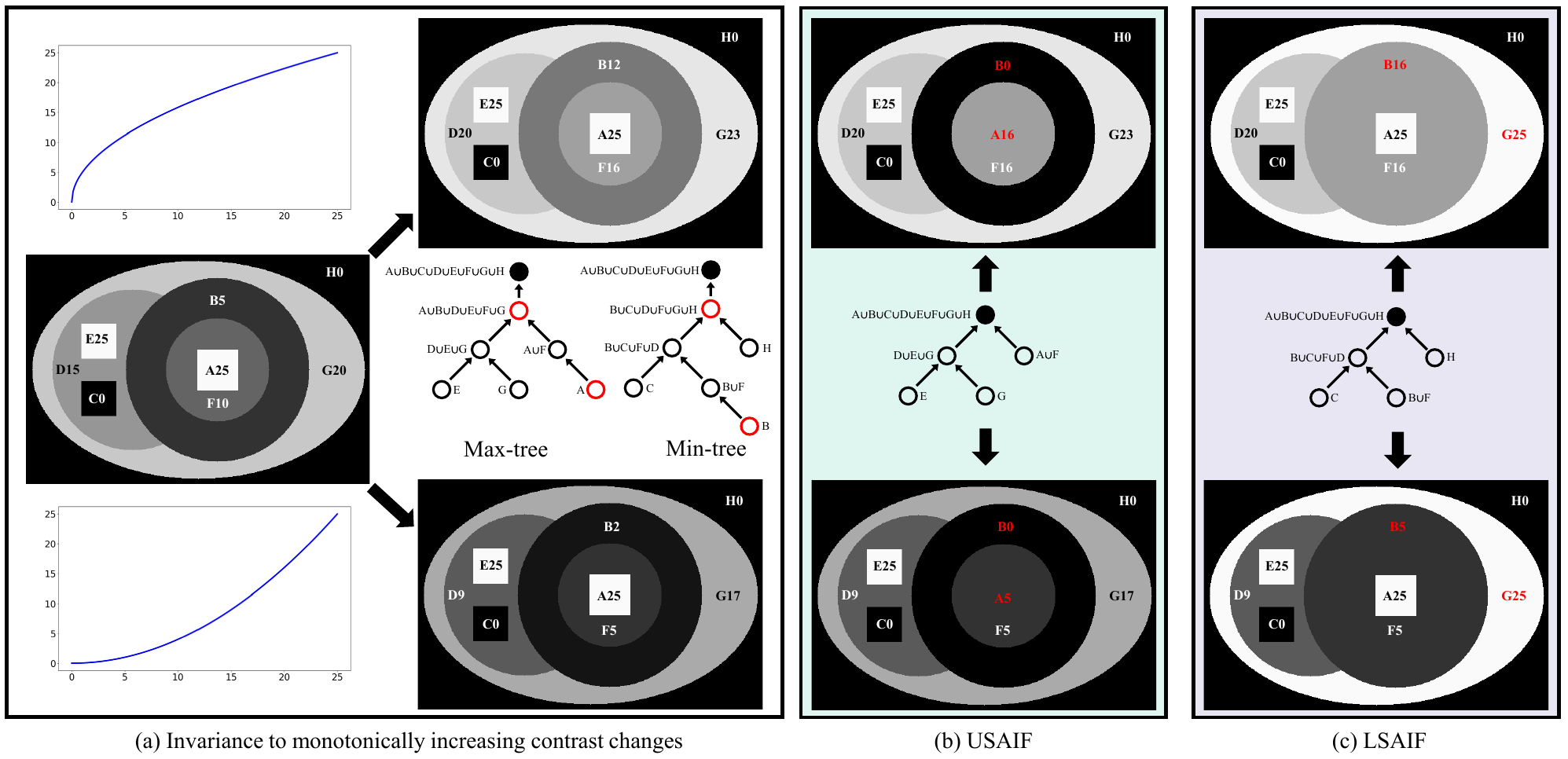}
\caption{An illustrative example of leveraging the contrast-invariance property (a) of Max/Min-tree in DSAIF. Applying monotonically increasing contrast changes before DSAIF increases the appearance diversity while preserving the same topological structure as the original images. }
%\caption{A synthetic example explaining that the monotone increasing intensity transformations can generate images with diverse intensity but similar structure. Using monotone increasing intensity transformations in pixel space does not change Max-tree/Min-tree structure, which keeps connected components unchanged under different transformations. However, the representation pixels of connected components are different under various transformations, so the reconstructed images are structure-invariant and intensity-variant. }
\label{fig:image2treeaug}
\end{figure*}

The topology of the tree encodes the topology of the image structure. The leaf nodes correspond to local regional maxima (\resp minima) in the Max-tree (\resp Min-tree). A node having more than one child signifies the fusion of two connected components, triggering a topological change of tree structure and thus image structure. A node having no siblings is topologically equivalent to its parent. Therefore, removing all nodes having no siblings does not change the topological structure of the image. This gives a simplified tree preserving topologically critical nodes. The filtered image reconstructed from the simplified tree has the same topological structure as the original image, but with different appearances. Such filter $\psi$ is called upper/lower structure-aware image filter denoted as USAIF and LSAIF for the use of Max-tree and Min-tree, respectively. Since the graylevel of the parent is smaller (\resp larger) than the graylevel of the current node in Max-tree (\resp Min-tree), the novel USAIF $\psi_M$ (\resp LSAIF $\psi_m$) actually belongs to the family of upper-leveling (\resp lower-leveling)~\citep{xu2015connected}. The filtered image by USAIF is no brighter than the original image, and satisfies the property: for any pair of neighboring points $(v_1, v_2): \psi_M(x)(v_1) > \psi_M(x)(v_2) \Rightarrow \psi_M(x)(v_1) \leq x(v_1)$. By duality, the filtered image by LSAIF is no darker than the original image and has the property: for any pair of neighboring points $(v_1, v_2): \psi_m(x)(v_1) > \psi_m(x)(v_2) \Rightarrow \psi_m(x)(v_2) \geq x(v_2)$. An illustrative example of the proposed dual structure-aware image filters (DSAIF) is given in Fig.~\ref{fig:image2tree}. 

The dual structure-aware image filterings USAIF and LSAIF preserve the same topological structure as the original image while generating diverse image appearances different from the original one. It is noteworthy that different from classical monotonically increasing contrast changes (\eg, Gamma correction) where pixels with the same graylevel have the same output graylevel, the proposed DSAIF may yield different output graylevels for the same input graylevel (see A and E in Fig.~\ref{fig:image2tree}(a)). Since small regions may be caused by noise, and do not contribute to the topological changes, we remove all nodes whose area is smaller than $\tau$ before performing DSAIF. The algorithm for the proposed strucutre-aware image filtering is given in Algorithm~\ref{alg:tree-component-preserving}.

As illustrated in Fig.~\ref{fig:image2tree}, an image can be viewed as a topological landscape with peaks and valleys. The topological structure of the landscape (\ie image) is well reflected by the structure of Max-tree and Min-tree, where the leaf nodes represent peaks and valleys, respectively. Medical objects of interest often have some prior topological structure (\eg, containing some peaks or valleys). The proposed DSAIF removes topologically equivalent nodes while preserving the critical ones whose merging triggers topological change. 
When the topological landscape has only one local minimum and one local maximum simultaneously, 
some regions with different gray levels may be merged into one in both USAIF and LSAIF. However, this is very rare in practice. Otherwise, either LSAIF or USAIF preserves the differentiated gray levels with the surrounding context. The use of both LSAIF and USAIF helps to effectively alleviate the confirmation bias problem during noisy pseudo label learning.

Since the Max-tree and Min-tree are invariant to monotonically increasing contrast changes, we further increase the appearance diversity while preserving the topological structure by applying some monotonically increasing contrast changes to the original image before performing DSAIF. Specifically, we use Gamma correction or monotonic Bézier Curve for each training image. For the Gamma correction augmentations, we independently random two Gamma values within $[0.5, 1.5]$ to generate two different views of the image. A B\'ezier curve is a parametric curve defined by a set of control points. In this paper, we use two end points ($P_{0}$ and $P_{3}$) and two control points ($P_{1}$ and $P_{2}$) to generate cubic Bézier curves $B(t)$:
\begin{equation}\label{Bezier}
B(t) = (1-t)^{3}P_{0} + 3(1-t)^{2}tP_{1}  + 3(1-t)t^{2}P_{2} + t^{3}P_{3}, ~t\in [0,1],
\end{equation}
where $t$ is a fractional value along the length of the line. We set $P_{0}=(-1,-1)$ and $P_{3}=(1,1)$ as fixed points. Then, we set $P_{1}=(-z, z)$ and $P_{2}=(z, -z)$, where $z \in [0,1]$. In each iteration, we randomly choose $z_1$ and $z_2$ from $\{0, 0.5, 0.75\} $ to generate two transform functions $B_{{1}}$ and $B_{{2}}$ applied to the input image. As illustrated in Fig.~\ref{fig:image2treeaug}, applying such monotonically increasing contrast changes to the image before performing DSAIF yields many different alternatives with diverse appearances while preserving the topological structure of the original image.

\subsection{Mutual Supervision on Dual Structure-Aware Filtered Images}

 \noindent \textbf{Network architecture:}
 The model consists of two networks $f_{\theta_{1}}$ and $f_{\theta_{2}}$ with the same network architecture but different parameter initializations $\theta_{1}$ and $\theta_{2}$. %The two networks have the same structure and their parameters $\theta_{1}$ and $\theta_{2}$  initiated differently. 
 For each image $x$, we apply the monotonically increasing contrast changes and the proposed DSAIF described in Sec.~\ref{subsec:tree-filter} to generate two different views $x_1$ and $x_2$ preserving the topological structure of the original image as the input for $f_{\theta_{1}}$ and $f_{\theta_{2}}$, respectively.
 
\noindent \textbf{Training objective:}
For each labeled image $x^l$, we adopt the cross-entropy loss $\ell_{ce}$ and dice loss $\ell_{dc}$ as the supervised loss $\mathcal{L}_s$ given by:
\begin{equation}\label{labeled supervise}
\mathcal{L}_s =  \ell_{ce}({p}_{1}^l, {y}^l) + \ell_{dc}({p}_{1}^l, {y}^l) + \ell_{ce}({p}_{2}^l, {y}^l) 
 + \ell_{dc}({p}_{2}^l, {y}^l),
\end{equation}
where ${p}_{1}^l$ and ${p}_{2}^l$ are the prediction output of the two networks, and $y^l$ is the corresponding label. For each unlabeled image $x^u$, we use the pseudo label obtained from one network to supervise the output of another one. The loss $\mathcal{L}_u$ for the unlabeled image $x^u$ is given by:
\begin{equation}\label{unlabeled supervise}
\mathcal{L}_u = \ell_{ce}({p}_{1}^u, \hat{y}_2) + \ell_{ce}({p}_{2}^u, \hat{y}_1),
\end{equation}
where $\hat{y}_1$ and $\hat{y}_2$ are pseudo labels obtained from $p_{1}^{u}$ and $p_{2}^{u}$, respectively. The overall training objective $\mathcal{L}$ is defined by:
\begin{equation} \label{eq:overallloss}
     \mathcal{L} =\mathcal{L}_s + \lambda \times \mathcal{L}_u, 
\end{equation}
 where $\lambda$ balances the two loss terms.
% Besides, We also simply adopt independently random rotation as geometry transformation to different views. All these transformations results in two diverse views of the same input image.
\begin{figure}[tp]
\vspace{-10pt}
\centering
\subfloat[Image]{
\begin{minipage}[b]{0.23\linewidth}
\includegraphics[width=1\linewidth]{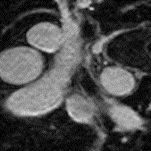}\vspace{3pt}
\includegraphics[width=1\linewidth]{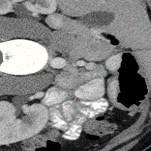}\vspace{3pt}
\includegraphics[width=1\linewidth]{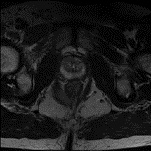}\vspace{3pt}
\end{minipage}}\hspace{-1mm}
\subfloat[Changed image]{
\begin{minipage}[b]{0.23\linewidth}
\includegraphics[width=1\linewidth]{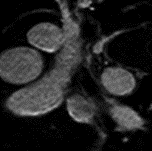}\vspace{3pt}
\includegraphics[width=1\linewidth]{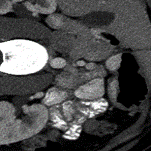}\vspace{3pt}
\includegraphics[width=1\linewidth]{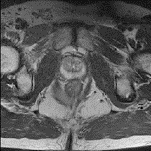}\vspace{3pt}

\end{minipage}}\hspace{-1mm}
\subfloat[USAIF]{
\begin{minipage}[b]{0.23\linewidth}
\includegraphics[width=1\linewidth]{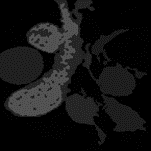}\vspace{3pt}
\includegraphics[width=1\linewidth]{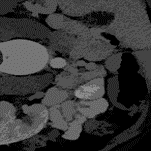}\vspace{3pt}
\includegraphics[width=1\linewidth]{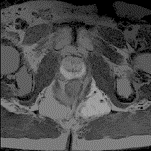}\vspace{3pt}
\end{minipage}}\hspace{-1mm}
\subfloat[LSAIF]{
\begin{minipage}[b]{0.23\linewidth}
\includegraphics[width=1\linewidth]{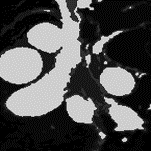}\vspace{3pt}
\includegraphics[width=1\linewidth]{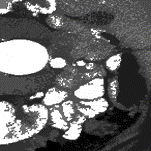}\vspace{3pt}
\includegraphics[width=1\linewidth]{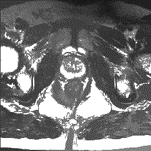}\vspace{3pt}

\end{minipage}}
\caption{Some qualitative results of DSAIF on LA dataset~\citep{xiong2021global} (first row), Pancreas-CT~\citep{clark2013cancer} (middle row), and PROMISE12~\citep{litjens2014evaluation} (bottom row). The changed images in (b) are obtained by applying monotonically increasing contrast change to the original ones.}
\label{fig:tree transform example}
\end{figure}

\section{Experiments}
\label{sec:experiments}
\subsection{Dataset and Evaluation Protocal}
Following some existing semi-supervised semantic segmentation methods, we mainly conduct experiments on the widely used 3D Left Atrium Segmentation MR Dataset (LA)~\citep{xiong2021global}, Pancreas-NIH~\citep{clark2013cancer}, and PROMISE12 dataset~\citep{litjens2014evaluation}.

\medskip
\noindent \textbf{LA Dataset:} 3D Left Atrial Segmentation Challenge dataset~\citep{xiong2021global} consists of 100 MRI scans. Following Wu \etal~\citep{wu2022mutual}, a fixed split is utilized, where 80 samples are designated for training and the remaining 20 samples are allocated for testing. 
% We conduct 3D segmentation experiments on Pancreas-CT dataset and LA dataset followed Wu \etal\citep{wu2022mutual} for fair comparisons. 

\medskip
\noindent \textbf{Pancreas-NIH Dataset:} Pancreas-NIH dataset~\citep{clark2013cancer} consists of 82 3D abdominal contrast-enhanced CT scans. Following the commonly-used data split in Luo~\citep{luo2021semi}, we take 62 samples for training and the rest 20 samples for testing. 
 % For image pre-processing, the voxel values were clipped to the range of [-125, 275] Hounsfield Units (HU), and the data was further resampled to an isotropic resolution of 1.0 × 1.0 × 1.0 mm in a similar way of ~\citep{zhou2019prior}.

\medskip
\noindent \textbf{PROMISE12 Dataset:} PROMISE12 dataset~\citep{litjens2014evaluation} consists of 50 transverse T2-weighted MRI scans. Following the data split in Liu \etal~\citep{liu2022semi}, there are 35, 5, and 10 scans for training, validation, and testing. Due to the low cross-slice resolution, PROMISE12 dataset are segmented in 2D (slice by slice)~\citep{liu2022semi}. 
% Each slice was resized to $256 \times 256$, and the intensity of pixels were normalized to the [0, 1] range. 

% \noindent \textbf{ACDC Dataset.} ACDC dataset~\citep{bernard2018deep} consists of 100 annotated MRI volumes. Following the data split in Luo \etal~\citep{luo2021efficient}, we take 70, 10, and 20 samples for training, validation, and testing, respectively. We conduct slice by slice 2D segmentation experiment on ACDC dataset. 

\medskip
\noindent \textbf{Evaluation protocol:} The proposed method is evaluated with four widely used metrics in semi-supervised medical image segmentation: Dice coefficient (Dice), Jaccard Index (JAC), the 95\% Hausdorff Distance (95HD), and the average surface distance (ASD).

\subsection{Implementation Details}
The SGD optimizer with a learning rate $10^{-2}$ and a weight decay factor $10^{-4}$ is used for all experiments. 
The loss weight $ \lambda$ in Eq.~\eqref{eq:overallloss} is set as a time-dependent Gaussian warming-up function \citep{laine2016temporal} using the same parameters as MC-Net+~\citep{wu2022mutual}. We adopt the V-Net (\resp U-Net) model as the backbone for 3D (\resp 2D) segmentation tasks following the same settings in MC-Net+~\citep{wu2022mutual} for fair comparisons. The area threshold parameter $\tau$ in DSAIF is set to 50 in 2D segmentation experiments and 100 in 3D segmentation experiments. All the experiments are conducted using the Pytorch framework with two NVIDIA GeForce RTX 3090 GPUs.

\begin{figure}[tp]
\vspace{-10pt}
\centering
\subfloat[UA-MT]{
\begin{minipage}[b]{0.155\linewidth}
\includegraphics[width=1\linewidth]{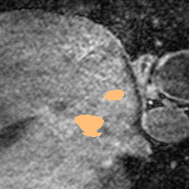}\vspace{3pt}
\includegraphics[width=1\linewidth]{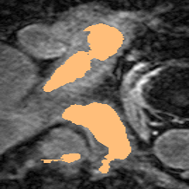}\vspace{3pt}
\includegraphics[width=1\linewidth]{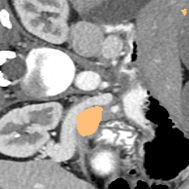}\vspace{3pt}
\includegraphics[width=1\linewidth]{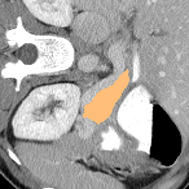}\vspace{3pt}
\includegraphics[width=1\linewidth]{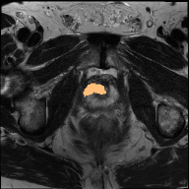}\vspace{3pt}
\includegraphics[width=1\linewidth]{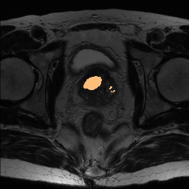}
\end{minipage}}\hspace{-2mm}
\subfloat[URPC]{
\begin{minipage}[b]{0.155\linewidth}
\includegraphics[width=1\linewidth]{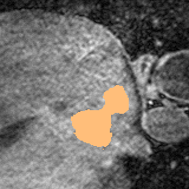}\vspace{3pt}
\includegraphics[width=1\linewidth]{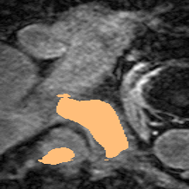}\vspace{3pt}
\includegraphics[width=1\linewidth]{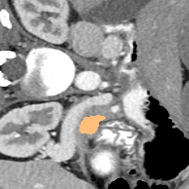}\vspace{3pt}
\includegraphics[width=1\linewidth]{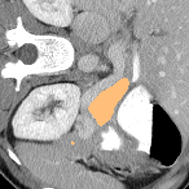}\vspace{3pt}
\includegraphics[width=1\linewidth]{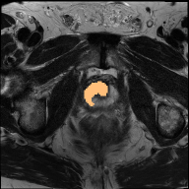}\vspace{3pt}
\includegraphics[width=1\linewidth]{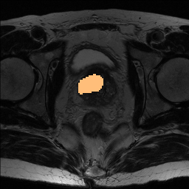}
\end{minipage}}\hspace{-2mm}
\subfloat[MC-Net+]{
\begin{minipage}[b]{0.155\linewidth}
\includegraphics[width=1\linewidth]{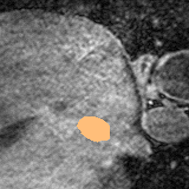}\vspace{3pt}
\includegraphics[width=1\linewidth]{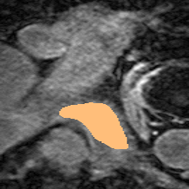}\vspace{3pt}
\includegraphics[width=1\linewidth]{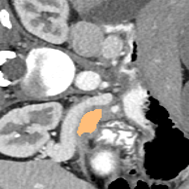}\vspace{3pt}
\includegraphics[width=1\linewidth]{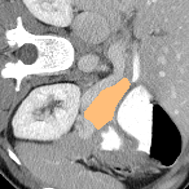}\vspace{3pt}
\includegraphics[width=1\linewidth]{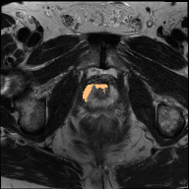}\vspace{3pt}
\includegraphics[width=1\linewidth]{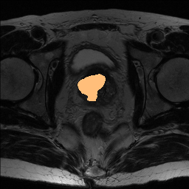}
\end{minipage}}\hspace{-2mm}
\subfloat[CPS]{
\begin{minipage}[b]{0.155\linewidth}
\includegraphics[width=1\linewidth]{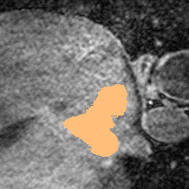}\vspace{3pt}
\includegraphics[width=1\linewidth]{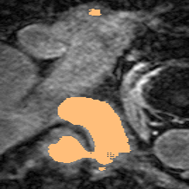}\vspace{3pt}
\includegraphics[width=1\linewidth]{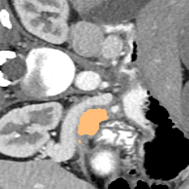}\vspace{3pt}
\includegraphics[width=1\linewidth]{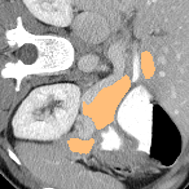}\vspace{3pt}
\includegraphics[width=1\linewidth]{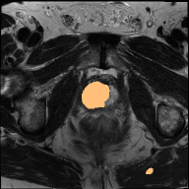}\vspace{3pt}
\includegraphics[width=1\linewidth]{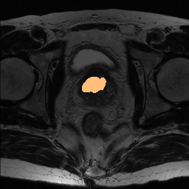}
\end{minipage}}\hspace{-2mm}
\subfloat[DSAIF]{
\begin{minipage}[b]{0.155\linewidth}
\includegraphics[width=1\linewidth]{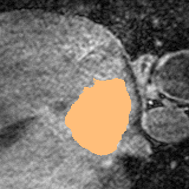}\vspace{3pt}
\includegraphics[width=1\linewidth]{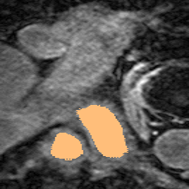}\vspace{3pt}
\includegraphics[width=1\linewidth]{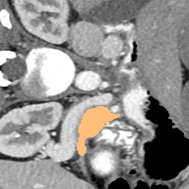}\vspace{3pt}
\includegraphics[width=1\linewidth]{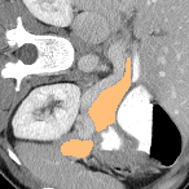}\vspace{3pt}
\includegraphics[width=1\linewidth]{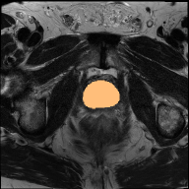}\vspace{3pt}
\includegraphics[width=1\linewidth]{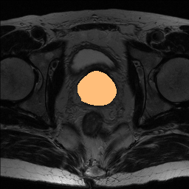}
\end{minipage}}\hspace{-2mm}
\subfloat[GT]{
\begin{minipage}[b]{0.155\linewidth}
\includegraphics[width=1\linewidth]{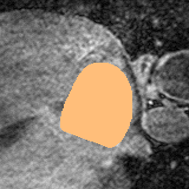}\vspace{3pt}
\includegraphics[width=1\linewidth]{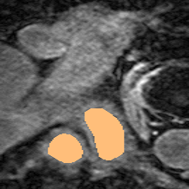}\vspace{3pt}
\includegraphics[width=1\linewidth]{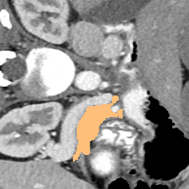}\vspace{3pt}
\includegraphics[width=1\linewidth]{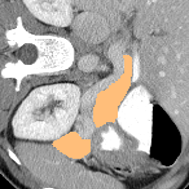}\vspace{3pt}
\includegraphics[width=1\linewidth]{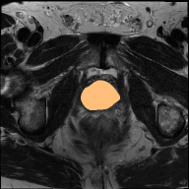}\vspace{3pt}
\includegraphics[width=1\linewidth]{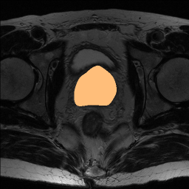}
\end{minipage}}
\caption{Some  qualitative segmentation results of  DSAIF on LA dataset~\citep{xiong2021global} (first two rows), Pancreas-CT dataset~\citep{clark2013cancer} (middle two rows), and PROMISE12 dataset~\citep{litjens2014evaluation} (bottom two rows).
}
\label{fig:vis}
\end{figure}

\subsection{Qualitative Results of DSAIF}
\label{subsec:qualitiveDSAIF}
Some qualitative results of the proposed DSAIF are shown in Fig.~\ref{fig:tree transform example}. Both USAIF and LSAIF generate images with diverse appearances while preserving the same topological structure as the original image. Inheriting from the property of connected filters, the proposed DSAIF does not create any new contours. It is also noteworthy that monotonically increasing contrast change map pixels with the same graylevel to the same output graylevel. Differently, the output of DSAIF does not only depend on the input graylevel, but also the image structure. As shown in the first row of Fig.~\ref{fig:tree transform example}, for similar input graylevels on different pixels, USAIF may output very different graylevels on these pixels. Yet, the topological image structure is preserved. 

%effectively transforms the grayscale values of pixels with similar intensities at various positions in (b) to significant differences values. 

%Furthermore, DSAIF maintains the fundamental structure of organs in the image and avoiding the introduction of any new contours.

\subsection{Comparative Results on Different Datasets}
\label{subsec:results}

% % \subsection{Visualization of results}
%  \begin{figure}[!tp]
%     \centering
%     \includegraphics[width=1\linewidth]{fig/experiments/vis.pdf}
%     \caption{Some qualitative illustrations on Pancreas-CT dataset~\citep{clark2013cancer}, LA dataset~\citep{xiong2021global}, ACDC dataset~\citep{bernard2018deep}, and PROMISE12 dataset~\citep{litjens2014evaluation} (from top to bottom). }
%     \label{fig:vis} 
% \end{figure}
Some qualitative segmentation results on the three datasets are shown in Fig.~\ref{fig:vis}, where we can observe that the proposed DSAIF achieves accurate segmentation results. 
We compare our proposed method with several state-of-the-art methods in the field of medical semi-supervised segmentation. \textbf{Among them, UA-MT~\citep{yu2019uncertainty}, CVRL~\citep{you2022momentum}, SS-Net~\citep{wu2022exploring}, SimCVD~\citep{you2022simcvd}, LLRU~\citep{adiga2022leveraging}, DUO-Net~\citep{peiris2021duo}, SCO-SSL~\citep{xu2021shadow}, BCP~\citep{bai2023bidirectional}, AC-MT~\citep{xu2023ambiguity}, AAU~\citep{adiga2024anatomically} also apply image-level variations.} 
% employ the consistency regularization strategy of image-level variation. 
The proposed DSAIF outperforms these image-level variations methods on three datasets, indicating the effectiveness of structure-aware image-level perturbations for semi-supervised medical image segmentation.

\begin{table}
  \centering
  \caption{Quantitative evaluation on the LA dataset~\citep{xiong2021global}.~\dag~represents the reproduced results based on the open-sourced implementation. We report the mean and standard deviation obtained over three runs.}
  \label{tab:LA}
  \setlength{\tabcolsep}{1mm}
   \resizebox{1\linewidth}{!}{
  \begin{tabular}{clcccc}
    \hline
    L/U &Method &Dice (\%)$\uparrow$ &JAC (\%)$\uparrow$ &95HD$\downarrow$ &ASD$\downarrow$ \\
    \hline
     80/0 &V-Net\dag &91.82 &84.92 &5.12 &1.71 \\
    \hline
    8/0 &V-Net\dag &80.75 &69.81 &15.11 &3.61 \\
    \hline
    % \multirow{10}*{8/72} &UA-MT~\citep{yu2019uncertainty}  &86.28 &76.11 &18.71 &4.63\\
    %  \multirow{10}*{(10\%)} &URPC~\citep{luo2021efficient}    &85.01 &74.36 &15.37 &3.96 \\
     % &GBDL~\citep{wang2022rethinking}       &88.40   &79.2  &\textbf{5.89}  &\textbf{1.60} \\
    \multirow{10}*{8/72}   &CVRL~\citep{you2022momentum}     &88.56   &78.89  &8.22  &2.81 \\
    \multirow{10}*{(10\%)}    &SS-NET~\citep{wu2022exploring}    &88.55      &79.62   &7.49   &1.90\\
       &SimCVD~\citep{you2022simcvd}      &89.03   &80.34  &8.34 &2.59 \\
       &LLRU~\citep{adiga2022leveraging}      &86.58  &-  &11.82 &- \\
       &MC-Net$+$~\citep{wu2022mutual} &88.96 &80.25 &7.93 &1.86\\
      &AC-MT~\citep{xu2023ambiguity} &89.12 &80.46  &11.05 &2.19  \\
      &BCP~\citep{bai2023bidirectional} &89.62  &81.31  &6.81  &1.76  \\
       &AAU~\citep{adiga2024anatomically} &86.58  &-  &11.82  &- \\

       \cmidrule(lr){2-6}
      &MC-Net~\citep{wu2021semi}\dag     &87.89$\pm$\scriptsize{0.51}   &78.58$\pm$\scriptsize{0.51}  &10.47$\pm$\scriptsize{1.98}  &2.23$\pm$\scriptsize{0.48} \\
      &MC-Net + DSAIF &\textbf{90.63$\pm$\scriptsize{0.34}} &\textbf{82.92$\pm$\scriptsize{0.45}} &\textbf{6.69$\pm$\scriptsize{0.97}} &\textbf{1.57$\pm$\scriptsize{0.26}} \\
    &CPS~\citep{chen2021semi}\dag     &86.79$\pm$\scriptsize{0.42}   & 77.05$\pm$\scriptsize{0.56}  &14.19$\pm$\scriptsize{2.08}  &4.25$\pm$\scriptsize{0.40} \\
      &CPS + DSAIF &90.20$\pm$\scriptsize{0.14} &82.22$\pm$\scriptsize{0.23} &6.72$\pm$\scriptsize{0.19} &1.77$\pm$\scriptsize{0.12} \\
    \hline
    16/0 &V-Net\dag    &88.41 &79.43 &10.05 &2.40 \\
    \hline
   % &UA-MT~\citep{yu2019uncertainty}   &88.74 &79.94 &8.39 &2.32\\
   %    &URPC~\citep{luo2021efficient}    &88.74 &79.93 &12.73 &3.66 \\
      % &GBDL~\citep{wang2022rethinking}      &89.40   &82.20  &\textbf{4.03}  &\textbf{1.48} \\
       \multirow{9}*{16/64} &CVRL~\citep{you2022momentum}      &90.45   &83.02 &6.56  &1.81 \\
     \multirow{9}*{(20\%)}  &SimCVD~\citep{you2022simcvd}     &90.85   &83.80  &6.03 &1.86 \\
       &LLRU~\citep{adiga2022leveraging}      &88.60  &-  &7.61 &- \\
      &MC-Net$+$~\citep{wu2022mutual} &91.07&83.67 &5.84 &1.67\\
      &MCF~\citep{wang2023mcf} &88.71 &80.41 &6.32 &1.90\\
            &AC-MT~\citep{xu2023ambiguity} &90.31 &82.43  &6.21 &1.76 \\
      &BCP~\citep{bai2023bidirectional}\dag &90.03 &82.35 &6.17  &1.68  \\
    &AAU~\citep{adiga2024anatomically} &88.60  &-  &7.61  &- \\

      \cmidrule(lr){2-6}

     &MC-Net~\citep{wu2021semi}\dag     &90.21$\pm$\scriptsize{0.34}   &82.24$\pm$\scriptsize{0.63}  &6.78$\pm$\scriptsize{0.87}  &1.70$\pm$\scriptsize{0.31} \\
      &MC-Net + DSAIF &\textbf{91.63$\pm$\scriptsize{0.26}} &\textbf{84.61$\pm$\scriptsize{0.42}} &\textbf{5.33$\pm$\scriptsize{0.58}} &\textbf{1.35$\pm$\scriptsize{0.19}} \\
     &CPS~\citep{chen2021semi}\dag      &90.07$\pm$\scriptsize{0.30}  &82.08$\pm$\scriptsize{0.49}  &6.81$\pm$\scriptsize{0.41}  &1.99$\pm$\scriptsize{0.21} \\
     &CPS + DSAIF &91.20$\pm$\scriptsize{0.21} &83.90$\pm$\scriptsize{0.35} &5.49$\pm$\scriptsize{0.37} &1.57$\pm$\scriptsize{0.08}\\
    \hline
   \end{tabular} 
    }
\end{table}

\medskip
\noindent\textbf{Results on LA Dataset:} 
Tab.~\ref{tab:LA} depicts the quantitative evaluation of the LA dataset. The proposed method achieves consistent improvement in terms of all four metrics compared with other state-of-the-art methods, achieving 90.63\% and 91.63\% Dice coefficient using 10\% and 20\% labeled data, respectively. 
Using 20\% labeled data achieves $\sim$99.8\% Dice performance of using full set of labeled data. Under the setting of using 10\% labeled images, the proposed method outperforms the baseline CPS by 3.41\% Dice coefficient and 5.17\% Jaccard index. 
% Compared with the state-of-the-art method SimCVD~\citep{you2022simcvd}, the proposed method achieves an improvement of 1.17\% Dice coefficient and 1.88\% Jaccard index. 
% Using 20\% labeled data achieves $\sim$99.3\% Dice performance of using full set of labeled data.

\medskip
\noindent\textbf{Results on Pancreas-NIH Dataset:} The quantitative results on the Pancreas-CT dataset are shown in Tab.~\ref{table:resluts on Pancreas-CT}. 
%Many works have only conducted experiments under the setting of 20\% labeled ratio on the Pancreas-CT dataset. 
Under the setting of using 10\% labeled data, the proposed method significantly improves the baseline CPS~\citep{chen2021semi} by 5.96\% Dice coefficient and 7.45\% Jaccard index, and significantly outperforms the other state-of-the-art methods. 
%In particular, the proposed method outperforms the state-of-the-art method MC-Net+~\citep{wu2022mutual} by 5.96\% Dice coefficient and 7.45\% Jaccard index. 
% When using 20\% labeled data, compared with state-of-the-art method MC-Net+~\citep{wu2022mutual}, the Dice coefficient (\resp Jaccard Index) is improved from 80.59\% (\resp 68.08\%) to 82.66\% (\resp 70.75\% ). 
Besides, the proposed method using 20\% labeled data achieves $\sim$99.9\% performance (in Dice) of using full set of labeled data. Using 20\% labeled data under the CPS~\citep{chen2021semi} baseline, the best result among our three experiments is 82.90 Dice, 71.10 JAC, and 1.60 ASD, which is comparable to the results of BCP~\citep{bai2023bidirectional}.

%We also confirm the efficacy of our approach by visual analysis using test cases in Fig.~\ref{fig:vis}.

\begin{table}
  \centering
  \caption{Quantitative evaluation on the Pancreas-NIH dataset~\citep{clark2013cancer}. ~\dag~represents the reproduced results based on the open-sourced implementation. We report the mean and standard deviation obtained over three runs.}
  \label{table:resluts on Pancreas-CT}
  \setlength{\tabcolsep}{1mm}
\resizebox{1.0\linewidth}{!}{
  \begin{tabular}{clcccc}
    \hline
    L/U &Method &Dice (\%)$\uparrow$ &JAC (\%)$\uparrow$ &95HD$\downarrow$ &ASD$\downarrow$ \\
    \hline
    62/0 &V-Net\dag  &82.68 &71.05 &5.19 &1.41 \\
    \hline
    6/0 &V-Net\dag &68.69 &55.03 &13.47 &3.63 \\
    \hline
     \multirow{6}*{6/56} &UA-MT~\citep{yu2019uncertainty}  &66.44 &52.02  &17.04 &3.03 \\
     \multirow{6}*{(10\%)} &URPC~\citep{luo2021efficient}   &73.53 &59.44 &22.57 &7.85\\
       &DTC\citep{luo2021semi}    &66.58 &51.79  &15.46  &4.16 \\
      &MC-Net$+$~\citep{wu2022mutual}  &74.01 &60.02 &12.59 &3.34\\
      &BCP~\citep{bai2023bidirectional}\dag &75.57 &61.35 &27.29  &8.16  \\
        &MLRP~\citep{su2024mutual}  &75.93 &62.12 &9.07 &\textbf{1.54} \\
      
      \cmidrule(lr){2-6}
      &MC-Net~\citep{wu2021semi}\dag     &69.47$\pm$\scriptsize{0.82}   &55.28$\pm$\scriptsize{1.15}  &20.74$\pm$\scriptsize{1.98}  &5.41$\pm$\scriptsize{0.67} \\
      &MC-Net + DSAIF &70.49$\pm$\scriptsize{0.84} &56.63$\pm$\scriptsize{1.27} &13.03$\pm$\scriptsize{2.12} &2.48$\pm$\scriptsize{0.84} \\
     &CPS~\citep{chen2021semi}\dag   &74.95$\pm$\scriptsize{0.91}  &60.86$\pm$\scriptsize{1.06}  &13.49$\pm$\scriptsize{1.87}  &4.59$\pm$\scriptsize{0.77} \\
     &CPS + DSAIF  &\textbf{80.91$\pm$\scriptsize{0.70}} &\textbf{68.31$\pm$\scriptsize{0.92}} &\textbf{7.67$\pm$\scriptsize{1.63}} &2.18$\pm$\scriptsize{0.25} \\
    \hline
    12/0 &V-Net\dag & 76.91 &63.86  &8.16  &2.07 \\
    \hline
    \multirow{8}*{12/50} &UA-MT~\citep{yu2019uncertainty}    &76.10 &62.62 &10.84 &2.43\\
     \multirow{8}*{(20\%)} &URPC~\citep{luo2021efficient}   &80.02 &67.30 &8.51 &1.98 \\
      &DTC\citep{luo2021semi}   &76.27  &62.82  &8.70  &2.20 \\
      &CVRL\citep{you2022momentum}       &76.68   &61.16  &8.24  &3.19 \\
       &SimCVD \citep{you2022simcvd}      &75.39   &61.56  &9.84 &2.33 \\
      &MC-Net$+$~\citep{wu2022mutual} &80.59 &68.08 &6.47 &1.74\\
      &MCF~\citep{wang2023mcf} &75.00 &61.27 &11.59 &3.27\\
      &BCP~\citep{bai2023bidirectional} &\textbf{82.91} &\textbf{70.79} &\textbf{6.43}  &2.25  \\
      &MLRP~\citep{su2024mutual}  &81.53 &69.35 &6.81 &\textbf{1.33 }\\

      \cmidrule(lr){2-6}
    &MC-Net~\citep{wu2021semi}\dag     &78.26$\pm$\scriptsize{0.35}   &65.12$\pm$\scriptsize{0.42}  &11.90$\pm$\scriptsize{2.04}  & 3.25$\pm$\scriptsize{0.83} \\
    &MC-Net + DSAIF  &79.13$\pm$\scriptsize{0.42} &66.23$\pm$\scriptsize{0.38} &7.77$\pm$\scriptsize{1.65} &1.85$\pm$\scriptsize{0.43} \\
        &CPS~\citep{chen2021semi}\dag  &79.63$\pm$\scriptsize{0.14} & 66.77$\pm$\scriptsize{0.19}    & 8.77$\pm$\scriptsize{1.46}  & 2.33$\pm$\scriptsize{0.62}  \\
     &CPS + DSAIF &82.66$\pm$\scriptsize{0.32}  &70.75$\pm$\scriptsize{0.44}  &7.11$\pm$\scriptsize{1.01}  &1.75$\pm$\scriptsize{0.22}  \\
    \hline
  \end{tabular}
   }
\end{table}

\medskip
\noindent \textbf{Results on PROMISE12 Dataset:} On the PROMISE12 dataset, the proposed method achieves even more significant improvements over the other methods. In particular, as depicted in 
Tab.~\ref{tab:PROMISE12}, the proposed method outperforms the CPS baseline by 18.27\% Dice and 21.67\% JAC (\resp, 11.93\% Dice and 13.35\% JAC) under the setting of using 10\% (\resp 20\%) labeled data. 
% Compared with state-of-the-art method SCP-Net~\citep{zhang2023self} on this dataset, the proposed method achieves an improvement of 15.61\% (\resp 6.93\%) Dice coefficient when using 10\% (\resp 20\%) labeled data. 
The more significant improvement on this dataset is probably because that the image variance within the dataset is more prominent, further demonstrating the effectiveness of the proposed DSAIF using structure information for semi-supervised medical image segmentation. 
Moreover, the proposed method using 20\% labeled data achieves $\sim$99.2\% performance (in Dice) of using full set of labeled data.
\begin{table}
  \centering
  \caption{Quantitative evaluation on the PROMISE12 dataset~\citep{litjens2014evaluation}. ~\dag~represents the reproduced results based on the open-sourced implementation. We report the mean and standard deviation obtained over three runs.}
  \label{tab:PROMISE12}
  \setlength{\tabcolsep}{1mm}
    \resizebox{1.0\linewidth}{!}{
  \begin{tabular}{clcccc}
    \hline
    L/U &Method &Dice (\%)$\uparrow$ &JAC (\%)$\uparrow$ &95HD$\downarrow$ &ASD$\downarrow$ \\
    \hline
   35/0 &U-Net\dag &84.63  &73.73  &3.63  &1.27  \\
    \hline
    4/0 &U-Net\dag    &51.72  &39.02  &43.30  &11.81 \\
     \hline
    \multirow{6}*{4/31} &UA-MT~\citep{yu2019uncertainty}\dag &49.81  &37.52  &74.03  &15.95   \\
     \multirow{6}*{(10\%)} &URPC~\citep{luo2021efficient}\dag      &51.70 &38.30 &75.19 & 22.32 \\
      &SCO-SSL~\citep{xu2021shadow}\dag &62.34  &48.04 &33.79  &9.39   \\
      &DUO-Net~\citep{peiris2021duo}\dag  &45.02  &32.64    &69.36   &29.62  \\
      &SALCC~\citep{liu2022semi}\dag       &63.85 &48.07  &40.68  &7.50 \\
      &MC-Net$+$~\citep{wu2022mutual}\dag    &55.40  &42.01  &21.79  & 5.51 \\
      &SCP-Net~\citep{zhang2023self} & 66.21  &-  &-  &11.56 \\
      &BCP~\citep{bai2023bidirectional}\dag &77.93 &64.35 &27.39  &6.65  \\

      \cmidrule(lr){2-6}
      &MC-Net~\citep{wu2021semi}\dag     &54.26$\pm$\scriptsize{4.29}   &41.44$\pm$\scriptsize{2.89}  &35.82$\pm$\scriptsize{4.63}  &10.48$\pm$\scriptsize{2.31} \\
      &MC-Net + DSAIF  &78.26$\pm$\scriptsize{1.02} &65.22$\pm$\scriptsize{0.86} &11.26$\pm$\scriptsize{1.12} &4.58$\pm$\scriptsize{0.68}\\
      &CPS~\citep{chen2021semi}\dag       &63.55$\pm$\scriptsize{2.50}  &48.08$\pm$\scriptsize{1.76}  &40.50$\pm$\scriptsize{2.66} &12.20$\pm$\scriptsize{1.42} \\
      &CPS + DSAIF     &\textbf{81.82$\pm$\scriptsize{0.18}} &\textbf{69.75$\pm$\scriptsize{0.27}} &\textbf{5.14$\pm$\scriptsize{0.68}} &\textbf{1.42$\pm$\scriptsize{0.11}} \\
    \hline
    7/0 &U-Net\dag &65.63  &52.44  &12.36  &2.98  \\
     \hline
    \multirow{6}*{7/28}  &UA-MT~\citep{yu2019uncertainty}    &61.55  &-  &-  &13.94 \\
     \multirow{6}*{(20\%)} &URPC~\citep{luo2021efficient}     &61.55 &- &- &9.63 \\
     &SCO-SSL~\citep{xu2021shadow}\dag   &74.60  &61.72 &18.40  &5.02  \\
     &DUO-Net~\citep{peiris2021duo}\dag  &7018  &5495    &22.14   &5.49  \\
      &SALCC~\citep{liu2022semi}     &70.30  &-  &-  & 4.69\\
      &MC-Net$+$~\citep{wu2022mutual}\dag      &66.91  &55.26  &12.74  &2.41 \\
      &SCP-Net~\citep{zhang2023self} &77.06  &-  &-  &3.52 \\
      &BCP~\citep{bai2023bidirectional}\dag &80.26 &67.46 &9.67  &4.87  \\

      \cmidrule(lr){2-6}
      &MC-Net~\citep{wu2021semi}\dag     &71.70$\pm$\scriptsize{1.26}   &59.71 $\pm$\scriptsize{1.98}  &19.48$\pm$\scriptsize{2.02}  &8.38$\pm$\scriptsize{1.34} \\
      &MC-Net + DSAIF &79.56$\pm$\scriptsize{1.67} &67.87$\pm$\scriptsize{2.12} &15.32$\pm$\scriptsize{2.48} &6.17$\pm$\scriptsize{1.54} \\
      &CPS~\citep{chen2021semi}\dag      &72.06$\pm$\scriptsize{2.73}  &59.53$\pm$\scriptsize{2.21}  &9.55$\pm$\scriptsize{1.72}  &1.71$\pm$\scriptsize{0.05} \\
      &CPS + DSAIF      &\textbf{83.99$\pm$\scriptsize{0.81}} &\textbf{72.88$\pm$\scriptsize{1.09}} &\textbf{5.04$\pm$\scriptsize{0.61}} &\textbf{1.24$\pm$\scriptsize{0.11}} \\
    \hline
  \end{tabular} 
   }
\end{table}

% \noindent \textbf{Results on ACDC Dataset.} To further evaluate the effectiveness of the proposed method, we conduct 2D segmentation experiments on the ACDC dataset. The quantitative results are shown in Table~\ref{tab:ACDC}. The proposed method scores the best in terms of all four metrics using both 10\% and 20\% labeled data. 
% \input{table/ACDC-new}

\subsection{Ablation Studies}

\begin{table}[t]
  \centering
  \caption{Abaltion study on LA dataset~\citep{xiong2021global} under 10\% labeled data using CPS~\cite{chen2021semi} as baseline. We report the mean and standard deviation obtained over three runs.}
  \label{tab:abaltion}
  \setlength{\tabcolsep}{1mm}
    \resizebox{1\linewidth}{!}{
  \begin{tabular}{ccc|cccc}
    % \hline
    % \multicolumn{3}{c|}{Setting}  &\multicolumn{4}{c}{LA} \\
    \hline
    CPS  &Aug &DSAIF  &Dice (\%) &JAC (\%) & 95HD  &ASD \\
    \hline
    \checkmark  &   &     &86.79$\pm$\scriptsize{0.42}   &77.05$\pm$\scriptsize{0.56}  &14.19$\pm$\scriptsize{2.08}    & 4.25$\pm$\scriptsize{0.40}\\
    \checkmark  &\checkmark   &   &87.20$\pm$\scriptsize{0.34}   &77.68$\pm$\scriptsize{0.48}   &12.58$\pm$\scriptsize{1.15}   &4.02$\pm$\scriptsize{0.53} \\
    \checkmark  &   &\checkmark    &88.17$\pm$\scriptsize{0.09}    &79.02$\pm$\scriptsize{0.14}    &11.26$\pm$\scriptsize{1.81}    & 3.02$\pm$\scriptsize{0.45} \\
    \checkmark  &\checkmark  &\checkmark   &\textbf{90.20$\pm$\scriptsize{0.14}}  & \textbf{82.22$\pm$\scriptsize{0.23}}  & \textbf{6.72$\pm$\scriptsize{0.19}}   & \textbf{1.77$\pm$\scriptsize{0.12}}\\
    \hline
  \end{tabular}
  }
\end{table}
We conduct ablation studies on LA dataset under the setting of using 10\% labeled data. As depicted in Tab.~\ref{tab:abaltion}, directly adopting monotonically increasing contrast changes and random rotation as data augmentations do not significantly improve the results (86.79\% to 87.20\% Dice). Applying the proposed DSAIF on the original images outperforms the baseline by 1.38\% Dice and 1.97\% Jaccard index. Besides, combining these data augmentations and DSAIF significantly boosts the segmentation results by 3.0\% Dice and 4.54\% Jaccard index. This demonstrates that the performance improvement is mainly brought by the proposed DSAIF. 

We also conduct an ablation study on the area threshold $\tau$ involved in the proposed DSAIF. As shown in Fig.~\ref{table:area}, different settings of $\tau$ slightly influence the results. Using too small values makes DSAIF sensitive to noise. Using too large values may filter out some important regions. Setting $\tau = 100$ gives the best result.
%  \begin{figure}[!tp]
%     \centering
%     \includegraphics[width=0.7\linewidth]{fig/experiments/area.pdf}
%     \caption{Ablation study on the area threshold $\tau$ involved in the proposed DSAIF on LA dataset~\citep{xiong2021global}. 
% }
%     \label{fig:area} 
% \end{figure}

 \begin{table}[t]
      \centering
         \caption{Ablation study on the area threshold $\tau$ involved in the proposed DSAIF on LA dataset~\citep{xiong2021global} under 10\% labeled data using CPS~\cite{chen2021semi} as baseline. }
\resizebox{0.8\linewidth}{!}{
     \begin{tabular}{c c c c c c}
     \toprule
        Threshold $\tau$  & 0 & 50 & 100 & 150 & 200  \\
    \midrule
        Dice (\%)  & 89.23 & 90.03 & 90.33 & 89.76 & 89.01 \\
    \bottomrule
     \end{tabular}
     }

     \label{table:area}
 \end{table}

%  \begin{figure}[!tp]
%     \centering
%     \includegraphics[width=0.9\linewidth]{fig/experiments/cutmix.pdf}
%     \caption{Ablation experiment on Cutmix augmentation. 
% }
%     \label{fig:cutmix} 
% \end{figure}

%  \begin{figure}[!tp]
%     \centering
%     \includegraphics[width=1\linewidth]{fig/experiments/boxmap.pdf}
%     \caption{Abaltion on the LA dataset using box diagram. }
%     \label{fig:boxmap} 
% \end{figure}

\subsection{Domain Generalization Results}

\begin{table}[]
\centering
\caption{Cross-dataset performance on prostate segmentation. We report the mean and standard deviation over three runs.}
\label{tab:Cross-domain}
\resizebox{1\linewidth}{!}{
\begin{tabular}{cc|ll|ll}
\hline
\multicolumn{2}{c|}{\multirow{2}{*}{Setting}}    & \multicolumn{2}{c|}{L/U: 4/31}    & \multicolumn{2}{c}{L/U: 7/28} \\ 
\cline{3-6}
\multicolumn{2}{c|}{}      & \multicolumn{1}{c}{Dice (\%)} &\multicolumn{1}{c|}{JAC (\%)}  & \multicolumn{1}{c}{Dice (\%)} 
& \multicolumn{1}{c}{JAC (\%)} \\
\hline

\multirow{3}{*}{\scriptsize{Site A}} &\multicolumn{1}{|l|}{\scriptsize{CPS (Baseline)}} &15.34$\pm$\scriptsize{1.32}  &9.89$\pm$\scriptsize{1.16}  &65.14$\pm$\scriptsize{2.08}  &52.50$\pm$\scriptsize{2.00}  \\ 
                &\multicolumn{1}{|l|}{\scriptsize{CPS + Aug}}  &28.80$\pm$\scriptsize{5.39}  &18.73$\pm$\scriptsize{4.02}  &70.43$\pm$\scriptsize{2.33}  &57.31$\pm$\scriptsize{1.29}    \\ 
                &\multicolumn{1}{|l|}{\scriptsize{CPS + DSAIF}}  &\textbf{40.36$\pm$\scriptsize{2.62} } &\textbf{27.89$\pm$\scriptsize{2.17} }  &\textbf{75.20$\pm$\scriptsize{0.33} } &\textbf{62.07$\pm$\scriptsize{0.61} }\\ 
\hline

\multirow{3}{*}{\scriptsize{Site B}} &\multicolumn{1}{|l|}{\scriptsize{CPS (Baseline)}}  &37.30$\pm$\scriptsize{6.86}  &25.84$\pm$\scriptsize{6.05}  &40.37$\pm$\scriptsize{ 6.91}  & 33.45$\pm$\scriptsize{4.21}  \\ 
                &\multicolumn{1}{|l|}{\scriptsize{CPS + Aug}}  &39.89$\pm$\scriptsize{1.84}  &28.10$\pm$\scriptsize{1.69}   &55.14$\pm$\scriptsize{2.97}  &41.03$\pm$\scriptsize{1.00}    \\ 
                &\multicolumn{1}{|l|}{\scriptsize{CPS + DSAIF}}  &\textbf{48.50$\pm$\scriptsize{4.01} } &\textbf{35.63$\pm$\scriptsize{3.92} } &\textbf{64.16$\pm$\scriptsize{2.89} } &\textbf{49.42$\pm$\scriptsize{2.95} }\\ 
\hline

% \multirow{3}{*}{Domain2} &\multicolumn{1}{|c|}{Baseline}  &3.10$\pm$\scriptsize{0.83}  &1.66$\pm$\scriptsize{0.52}  &9.26$\pm$\scriptsize{1.21}  &10.17$\pm$\scriptsize{5.65}  \\ 
%                 &\multicolumn{1}{|c|}{our1}  &20.58$\pm$\scriptsize{5.33}  &13.03$\pm$\scriptsize{3.77}   &25.89$\pm$\scriptsize{2.11}  &16.99$\pm$\scriptsize{1.67}    \\ 
%                 &\multicolumn{1}{|c|}{our2}  &21.77$\pm$\scriptsize{2.51}  &14.09$\pm$\scriptsize{1.88}   &38.48$\pm$\scriptsize{8.50}  &26.71$\pm$\scriptsize{6.68} \\ 
% \hline

\end{tabular}
}
\end{table}

%Incorporating shape and structural information about organs can enhance the generalization ability of the model. 
We conduct cross-domain experiments on prostate segmentation task in the semi-supervised setting to further verify the generalization performance of the proposed DSAIF. Under the setting of using 10\% (\resp 20\%) labeled data, we use 4 (\resp 7) labeled images and 31 (\resp 28) unlabeled images in PROMISE12 dataset to train the model, and test the model on 2 different data sources with distribution shift: Site A and B are from NCI-ISBI13 dataset~\citep{bloch2015nci}. As depicted in Tab.~\ref{tab:Cross-domain}, though the monotonically increasing contrast changes is helpful in domain generalization, the proposed DSAIF further significantly improves the baseline of using monotonically increasing contrast changes, demonstrating the effectiveness of DSAIF in domain generalization under semi-supervised setting. Specifically, under the setting of using 10\% labeled data, DSAIF achieves 11.56\% (\resp 9.16\%) Dice (\resp JAC) improvement on Site A and 8.61\% (\resp 7.53\%) Dice (\resp JAC) improvement on Site B. 
Under the setting of using 20\% labeled data, DSAIF achieves 4.77\% (\resp 4.76\%) Dice (\resp JAC) improvement  on Site A and a 9.02\% (\resp 8.39\%) Dice (\resp JAC) improvement on Site B. 
%It is noteworthy that the proposed DSAIF 

\subsection{Discussion}
\label{analysis-diversity}

% %%up down 
% \begin{figure}[tp]
% \vspace{-10pt}
% \centering
% \subfloat[]{
% \begin{minipage}[b]{0.8\linewidth}
% \includegraphics[width=1\linewidth]{fig/experiments/error_map.pdf}\vspace{3pt}
% \end{minipage}}\hspace{-2mm}
% \subfloat[]{
% \begin{minipage}[b]{0.8\linewidth}
% \includegraphics[width=1\linewidth]{fig/experiments/unlabeled_dice.pdf}\vspace{3pt}
% \end{minipage}}\hspace{-2mm}
% \caption{(a): ConBias in the training process. (b) The Dice coefficient between the ground-truth and the network outputs on unlabeled training images of PROMISE12 Dataset~\citep{litjens2014evaluation} at different iterations in the training process.
% }
% \label{fig:analysis}
% \end{figure}

% \begin{figure}[tp]
% \vspace{-10pt}
% \centering
% \subfloat[]{
% \begin{minipage}[b]{0.8\linewidth}
% \includegraphics[width=1\linewidth]{fig/experiments/error_map.pdf}\vspace{3pt}
% \includegraphics[width=1\linewidth]{fig/experiments/unlabeled_dice.pdf}
% \end{minipage}}\hspace{-2mm}
% \subfloat[]{
% \begin{minipage}[b]{0.8\linewidth}
% \includegraphics[width=1\linewidth]{fig/experiments/unlabeled_dice.pdf}\vspace{3pt}
% \end{minipage}}\hspace{-2mm}
% \caption{(a): ConBias in the training process. (b) The Dice coefficient between the ground-truth and the network outputs on unlabeled training images of PROMISE12 Dataset~\citep{litjens2014evaluation} at different iterations in the training process.
% }
% \label{fig:analysis}
% \end{figure}

 \begin{figure}[!tp]
    \centering
    \includegraphics[width=0.75\linewidth]{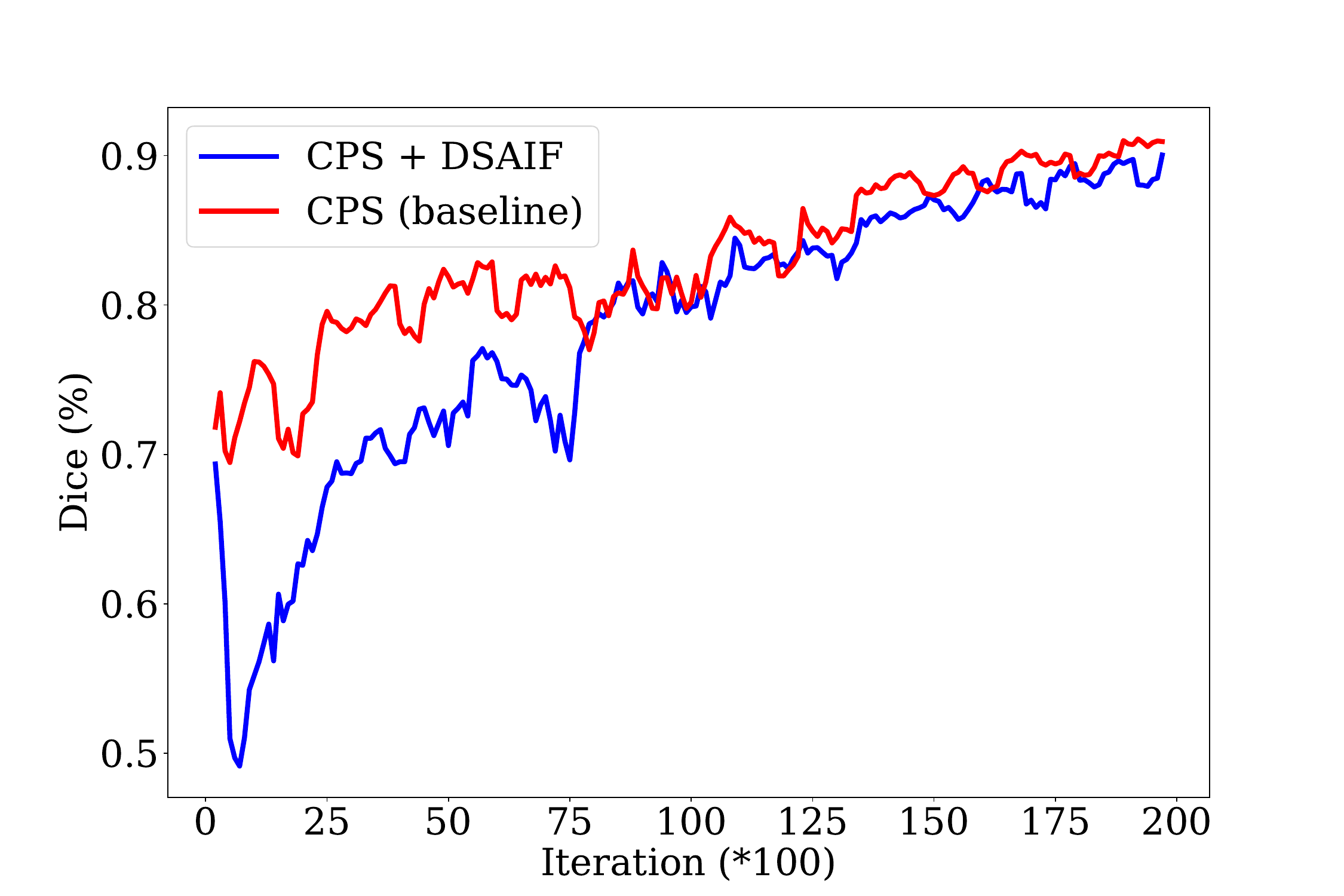}
    \caption{Dice score $D_{e}$ between erroneous predictions of two mutually supervised networks on unlabeled training images of PROMISE12 Dataset~\citep{litjens2014evaluation} during the training process.}
    \label{fig:error_prediction} 
\end{figure}

 \begin{figure}[!tp]
    \centering
    \includegraphics[width=0.75\linewidth]{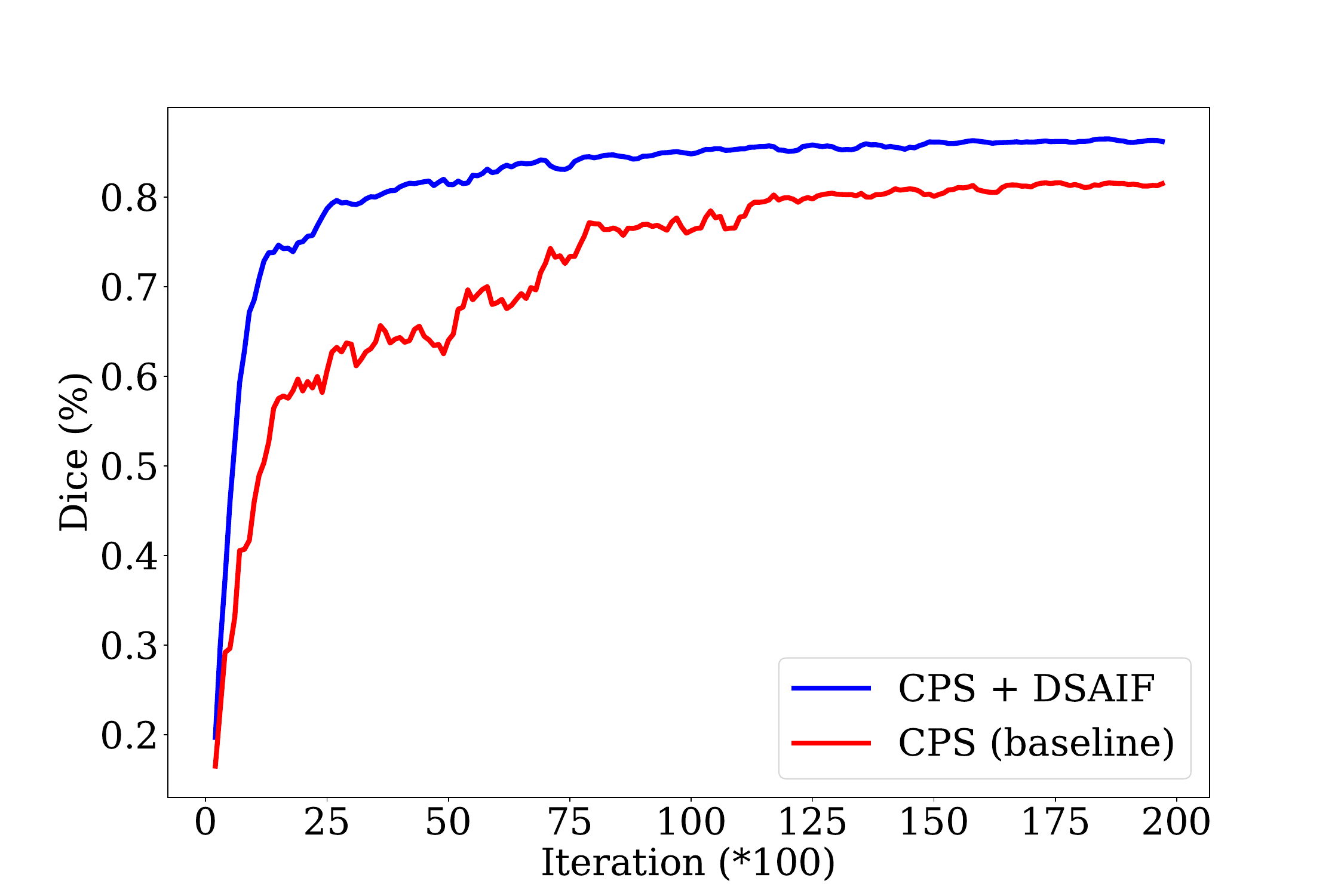}
    \caption{The Dice coefficient between the ground-truth and the network outputs on unlabeled training images of PROMISE12 Dataset~\citep{litjens2014evaluation} at different iterations in the training process.}
    \label{fig:unlabeled_dice} 
\end{figure}

% \begin{figure}
%     \centering
%     % \begin{minipage}
%     %     \includegraphics[width=0.45\linewidth]{fig/experiments/error_map.pdf}
%     % \end{minipage}
%     \begin{subfigure}{0.47\linewidth}
%          \includegraphics[width=1\linewidth]{fig/experiments/error_map.pdf}
%          \caption{}
%          \label{fig:error_prediction}
%     \end{subfigure}
%     \begin{subfigure}{0.47\linewidth}
%         \includegraphics[width=1\linewidth]{fig/experiments/unlabeled_dice.pdf}
%         \caption{}
%         \label{fig:unlabeled_dice}
%     \end{subfigure}
   
%     \caption{Dice Scores during training process on PROMISE12 Dataset~\citep{litjens2014evaluation}: (a) Erroneous predictions of two mutually supervised networks on unlabeled images, (b) Ground-truth and the network outputs on unlabeled images}

% % (a) is Dice score $D_{e}$ between erroneous predictions of two mutually supervised networks on unlabeled training images of PROMISE12 Dataset~\citep{litjens2014evaluation} during the training process. (b) is the Dice coefficient between the ground-truth and the network outputs on unlabeled training images of PROMISE12 Dataset~\citep{litjens2014evaluation} at different iterations in the training process. }
%     \label{fig:dice}
% \end{figure}

The pseudo-label-based semi-supervised medical image segmentation methods focus on generating  pseudo labels of high quality for unlabeled images. Since there are inevitable noisy labels in the pseudo labels for unlabeled images, it is critical to avoid the model overfitting to incorrect pseudo labels. Due to the absence of a clear supervision signal for the unlabeled image, when both networks make consistent incorrect predictions on some pixels, the mutual supervision between them may lead to a confirmation bias in the results. This makes the model overfit to noisy pseudo labels, yielding degenerated segmentation performance. Appropriate diversity between two networks' erroneous predictions helps to avoid such confirmation bias issue of overfitting to incorrect pseudo-labels. We define a quantitative metric $D_e$ to characterize such diversity of erroneous predictions on unlabeled training images between the two mutually supervised networks. For that, let $\mathcal{E}^1$ and $\mathcal{E}^2$ denote the set of pixels with incorrect prediction of the first and second network, respectively. We compute $D_e$ as the Dice score between $\mathcal{E}^1$ and $\mathcal{E}^2$ given by:
\begin{equation}\label{eq:diversity}
% \begin{aligned}
D_e = 2 \times |\mathcal{E}^1 \cap \mathcal{E}^2|/(|\mathcal{E}^1| + |\mathcal{E}^2|),
% \end{aligned}
\end{equation}
where $|\cdot|$ denotes the cardinality. The comparison of $D_e$ for the baseline model and the proposed method during the training process is depicted in Fig.~\ref{fig:error_prediction}. Thanks to the large appearance diversity between USAIF and LSAIF while preserving the same topological structure as the original image, the proposed DSAIF has less consensus on the erroneous predictions of the two mutually supervised networks. This helps to alleviate the confirmation bias issue of overfitting to noisy pseudo labels on unlabeled images, resulting in better pseudo labels of unlabeled images during the training process (see Fig.~\ref{fig:unlabeled_dice}). Therefore, the proposed DSAIF is effective in improving the performance of semi-supervised medical image segmentation. 

It is noteworthy that the tree of shapes~\citep{monasse2000fast}, known also as topographic maps~\citep{caselles1999topographic}, provides another way to convert the image into a structure-aware tree space. This leads to a \textbf{single} filtered image while preserving the topological structure of the original image. Yet, the proposed method requires generating \textbf{dual} images with very different appearances to decrease the consensus of erroneous predictions on unlabeled images. Therefore, we choose the Max/Min-tree representation in our DSAIF. However, it would be interesting to explore the use of tree of shapes for more structure-aware filters in semi-supervised medical image segmentation.  This is left for future work.

On the other hand, the prior knowledge about the topological structure of medical objects has also been explored in~\citep{clough2020topological,hu2019topology,hu2021topology,hu2022structure,gupta2022learning,singh2023topological} for medical image analysis. Most of them focus on designing topology-aware loss functions to incorporate the prior structure knowledge, helping to yield more plausible segmentation results. This is different from our DSAIF, which aims to generate dual images with different appearances while preserving the critical topological structure of the original image. This helps to cope with confirmation bias issue in semi-supervised medical image segmentation. It would also be interesting to combine DSAIF with these topological analysis tools in the future work. 

%It can be seen that our method uses tree space filtering to increase the structural differences between the input images of two networks, resulting in increased regional differences in prediction errors between the two networks during the training process, effectively avoiding formation bias. The Dice coefficient calculating by unlabeled images and ground truth of our method is significantly higher than baseline, which showns in Fig.~\ref{fig:unlabeled_dice}. 

A limitation of the current work is that the proposed DSAIF requires some extra time during the training process (but no extra runtime during inference). The implementation of DSAIF mainly involves the construction of Max/Min-tree, which can be achieved in quasi-linear time complexity with respect to the number of pixels/voxels~\citep{najman2006building, carlinet2014comparative}. Currently, we adopt CPU-based algorithm to build Max/Min-tree, which is not as efficient as GPU-based algorithm~\citep{blin2022max}. Yet, this GPU-based algorithm~\citep{blin2022max} does not support 3D images. In the future, we plan to explore the implementation of DSAIF with GPU to accelerate the training process. An alternative solution is to compute the DSAIF using offline strategy. 

%The Max-tree/Min-tree construction method in this paper is basic CPU-based algorithm, which limits the speed of network training. The
%first GPU algorithm to compute the max-tree has recently been proposed~\citep{blin2022max}. However, this GPU algorithm does not support 3D images. This prevents us from directly utilizing this algorithm to accelerate the training of our network. We plan to extend this GPU algorithm to 3D images in the future.

\section{Conclusion}
\label{sec:conclusion}
We propose a novel image-level variation method named dual structure-aware image filterings (DSAIF) for semi-supervised medical image segmentation. 
%in tree space that acts as an image-level variation  to cope with the confirmation bias issue in semi-supervised medical image segmentation.
Specifically, we leverage the dual Max-tree and Min-tree image representation, and remove all nodes having no siblings in the corresponding tree. This equals to remove all topologically equivalent regions while preserving topologically critical ones, resulting in two images with diverse appearances while having the same topological structure as the original image. 
% Applying the proposed DSAIF to mutually supervised networks decreases the consensus of their erroneous predictions on unlabeled images. This helps to alleviate the confirmation bias issue of overfitting to noisy pseudo labels of unlabeled images, and thus effectively improves the segmentation performance.
By incorporating the proposed DSAIF into mutually supervised networks, the consensus on erroneous predictions for unlabeled images is decreased. This helps to alleviate the confirmation bias issue, where models tend to overfit to noisy pseudo labels, thereby enhancing the performance of segmentation.
Extensive experimental results on three widely used benchmark datasets demonstrate that the proposed method significantly/consistently outperforms the state-of-the-art methods. 
%Changing the tree structure will result in different reconstructed images. How to design a reasonable tree filtering algorithm is a direction worth exploring in the future. 
In the future, we would like to explore DSAIF in more semi-supervised medical image segmentation frameworks, and using tree of shapes for more structure-aware filters. Combing DSAIF with other topological analysis tools is also an interesting direction to explore. 

%In this paper, we only add DSAIF to the CPS framework to verify its effectiveness, we will add DSAIF to more semi supervised frameworks in the future. Furthermore, we plan to generate Max-tree/Min-tree using GPUs to improve computational efficiency.

% \section*{Acknowledgments}
% Acknowledgments should be inserted at the end of the paper, before the
% references, not as a footnote to the title. Use the unnumbered
% Acknowledgements Head style for the Acknowledgments heading.

% \section*{References}

% Please ensure that every reference cited in the text is also present in
% the reference list (and vice versa).

% \section*{\itshape Reference style}

% Text: All citations in the text should refer to:
% \begin{enumerate}
% \item Single author: the author's name (without initials, unless there
% is ambiguity) and the year of publication;
% \item Two authors: both authors' names and the year of publication;
% \item Three or more authors: first author's name followed by `et al.'
% and the year of publication.
% \end{enumerate}
% Citations may be made directly (or parenthetically). Groups of
% references should be listed first alphabetically, then chronologically.

%%Harvard
\bibliographystyle{model2-names.bst}\biboptions{authoryear}
\bibliography{medima-main}

\clearpage
\setcounter{page}{1}
\section*{Supplementary Material on pipeline of DSAIF based on MC-Net~\cite{wu2022mutual}}

%This supplementary material provides a more detailed introduction to the submitted manuscript from the following aspects. \ref{sec:pipeline_suppl}. The pipelines of DSAIF based on MC-Net~\citep{wu2021semi}. \ref{sec:tree_construction_suppl}.The detailed implementation of Max-tree and Min-tree. \ref{sec:contrast_suppl}.Why Max-Tree and Min-tree are invariant to monotonically increasing contrast changes. \ref{sec:increasing_parameter_suppl}.Ablation studies for hyper-parameters in monotonically increasing contrast changes.

%\section{ The pipelines of DSAIF based on MC-Net.}
%\label{sec:pipeline_suppl}
% We incorporate the proposed DSAIF into MCNet~\citep{wu2021semi}. The pipeline of the proposed framework based on CPS~\citep{chen2021semi} is depicted in Fig.~\ref{fig:pipeline_cps_suppl}. CPS~\citep{chen2021semi} enforces consistency between two segmentation networks perturbed with distinct initializations while processing the same input image and employ the pseudo one-hot label map from one network to supervise the other network. DSAIF enables both networks to probabilistically receive images generated by either filtered Max-tree or filtered Min-tree in every iteration. When one network takes in an image generated using filtered Max-tree, the other network processes the same image generated with filtered Min-tree, and vice versa. 

\begin{figure}[h]
\centering
\includegraphics[width=0.5\textwidth]{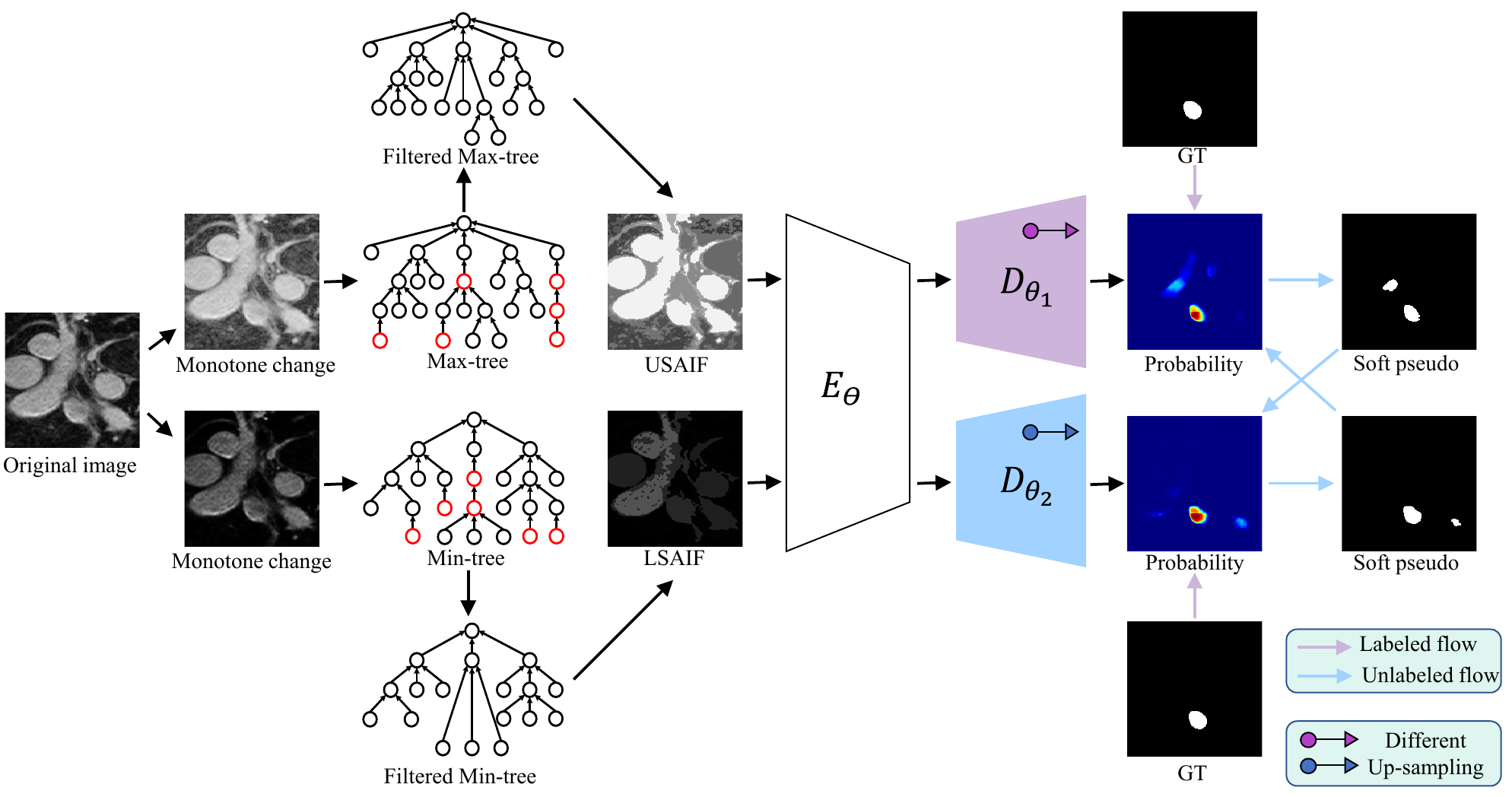}
\caption{The pipeline of the proposed DSAIF framework  using mutual supervision of MC-Net~\citep{wu2021semi} as the model-level variations. The pipeline composed of image-level variations and model-level variations on images. We propose novel dual structure-aware image filterings (DSAIF) based on Max/Min-tree representation as the image-level variations. }
\label{fig:pipeline_mcnet_suppl}
\end{figure}

The pipeline of the proposed framework based on MC-Net~\citep{wu2021semi} is depicted in Fig.~\ref{fig:pipeline_mcnet_suppl}. MC-Net~\citep{wu2021semi} comprises one shared encoder and two different decoders with distinct up-sampling strategies. Different from CPS, MC-Net~\citep{wu2021semi} introduces a mutual consistency constraint between the probability output of one decoder and the soft pseudo labels of the other decoder.  DSAIF enables the shared encoder to receive both Max-tree and Min-tree filtered images. The two decoders probabilistically receive features of images generated by either filtered Max-tree or filtered Min-tree in every iteration. When one decoder takes in features of the image generated with filtered Max-tree, the other decoder processes features of the same image generated with filtered  Min-tree, and vice versa.

\end{document}